\documentclass[11pt,letterpaper]{article}

\usepackage[hmargin=0.875in,vmargin=1in]{geometry}
\usepackage[utf8]{inputenc}
\usepackage[T1]{fontenc}
\usepackage{microtype}

\usepackage{amsmath,amssymb,amsthm,mathtools,bm}
\usepackage{amsfonts}
\usepackage[numbers,compress]{natbib}
\usepackage{graphicx}
\usepackage{booktabs}
\usepackage{array}
\usepackage{enumitem}
\usepackage{url}
\usepackage{xcolor}
\usepackage{tikz}
\usetikzlibrary{arrows.meta,positioning}
\usepackage{titletoc}
\usepackage[colorlinks=true,linkcolor=blue!60!black,citecolor=blue!60!black,urlcolor=blue!60!black]{hyperref}

\title{Flexible Doubly Robust Proximal Causal Inference \\ for Censored Outcomes}

\author{%
  Dhanajit Brahma \qquad Ruolin Meng \qquad Lawrence Carin \qquad Ricardo Henao\\[0.5ex]
  Duke University\\
  \texttt{\{dhanajit.brahma,ruolin.meng,lcarin,ricardo.henao\}@duke.edu}
}
\date{}

\newtheorem{theorem}{Theorem}
\newtheorem{proposition}{Proposition}
\newtheorem{lemma}{Lemma}

\theoremstyle{remark}

\newcommand{\E}{\mathbb{E}}
\newcommand{\En}{\mathbb{E}_n}
\newcommand{\Prob}{\mathbb{P}}
\newcommand{\cH}{\mathcal{H}}
\newcommand{\cT}{\mathcal{T}}
\newcommand{\cG}{\mathcal{G}}
\newcommand{\sigmoid}{\mathrm{sigmoid}}
\newcommand{\one}{\mathbf{1}}
\newcommand{\modelname}{\text{PSB}}
\newcommand{\modelnamePMMR}{\text{PSB-K}}
\newcommand{\modelnameNMMR}{\text{PSB-N}}
\newcommand{\modelfullform}{\textbf{P}roximal \textbf{S}urvival \textbf{B}ridge learning}
\newcommand{\Std}{\mathrm{Std}}

\newcommand{\indep}{\perp \!\!\!\! \perp}

\begin{document}

\maketitle

\begin{abstract}
Estimating causal effects on right-censored time-to-event outcomes is challenging under unmeasured confounding. Proximal causal inference addresses unmeasured confounding through proxy variables, but existing proximal survival methods rely largely on restrictive parametric bridge models. We introduce \modelfullform{} (\modelname{}), a flexible doubly robust proximal framework for censored outcomes. Building on existing proximal identification of treatment-specific survival curves, the framework yields outcome-bridge (OB), treatment-bridge (TB), and doubly robust (DR) estimators, with two flexible bridge learners: \modelnamePMMR{}, a kernel estimator with a low-rank time decomposition, and \modelnameNMMR{}, a neural estimator trained by a projected maximum-moment-restriction loss. Unlike existing flexible proximal learners for scalar outcomes, \modelname{} directly handles censoring and survival-curve targets. On a standard SEM benchmark, a proximal semi-synthetic ACTG study, and the real-world Framingham study, \modelnamePMMR{} performs strongest on SEM and ACTG, whereas \modelnameNMMR{} is the most stable on Framingham.  Additional results with modern causal survival baselines, weakened or invalid proxies, and heavier censoring assess robustness across challenging settings.
\end{abstract}

\section{Introduction}\label{sec:intro}
Accurate estimation of causal effects is essential in fields ranging from medicine to public health, particularly when assessing the impact of treatments or interventions on time-to-event outcomes, such as survival or disease recurrence \citep{rosenbaum2002observational,hernan2010causal,imbens2015causal}.
Causal survival analysis, which integrates causal inference with survival analysis \citep{kalbfleisch2002statistical}, provides the necessary framework to evaluate these impacts while formally accounting for the complexities of time and event occurrence \citep{denz2023comparison}.
However, drawing credible causal inferences from observational data is frequently hindered by the presence of {\em unmeasured confounding}.
Standard causal methods rely on the restrictive assumption of {\em conditional exchangeability} (or no unobserved confounding), asserting that all variables influencing both treatment and outcome have been measured and adjusted \citep{pearl2016causal}.
In real-world settings, unmeasured factors, often latent or poorly quantified, remain a persistent source of hidden bias, invalidating causal claims made solely on the basis of observed covariates \citep{rosenbaum1984association}.
This skepticism about unobserved confounding requires methods that can function with imperfect measurements.

The recently proposed {\em proximal causal inference} (PCI) framework offers a powerful and promising solution by formally identifying causal effects even when unmeasured confounding is present by strategically leveraging a pair of proxy variables, namely a {\em treatment proxy} and an {\em outcome proxy}, to serve as ``negative controls'' that capture the influence of the unmeasured confounder on the causal pathway \citep{kuroki2014measurement,miao2018identifying}.
This approach converts a non-identifiable problem into an identifiable one by capitalizing on the residual confounding structure contained within the proxy measurements.

Despite its theoretical promise, applying PCI to time-to-event data presents unique computational and inferential challenges.
First, survival analysis introduces the complexity of {\em censoring}, where the true event time is only partially observed, requiring specialized techniques to adjust for this form of missing data \citep{robins2000correcting}.
Second, nonparametric identification in the PCI framework requires estimating complex, often non-linear, confounding (outcome and treatment) bridge functions by solving an ill-posed integral equation \citep{tchetgen2020introduction}.
Relying on traditional, inflexible models for these functions can lead to significant specification bias.
Consequently, the development of robust and flexible estimation methods, such as doubly robust (DR) estimators \citep{ghassami2022minimax,wudoubly2024,cui2024semiparametric} integrated with deep neural networks \citep{xu2021deep,kompa2022deep,meng2025coupling}, is essential to ensure that causal estimates are consistent, efficient, and resilient to model misspecification when dealing with censoring.
However, most of the existing literature focuses on continuous or binary outcomes.

In this work, we address the largely understudied problem of PCI for time-to-event outcomes and
develop a doubly robust proximal framework for censored survival outcomes built around flexible bridge learning. Prior proximal survival work~\citep{ying2022proximal} uses low-dimensional parametric bridge models. Related work for censored outcomes~\citep{hu2025proximal} targets a scalar summary (the mean survival time), while prior neural and kernel proximal learners target scalar outcomes rather than censored survival curves. Our framework, \modelfullform{} (\modelname{}), closes this gap by combining survival-specific proximal identification with two flexible outcome-bridge learners and a targeted treatment-bridge stage, namely, \modelnamePMMR{} provides a regularized kernel solution that is effective in smoother regimes, while \modelnameNMMR{} replaces the kernel bridge representation with a neural network designed for harder misspecification through a more flexible nonlinear proxy structure. Both are then used to obtain outcome-bridge (OB), treatment-bridge (TB), and doubly robust (DR) estimators.
Our contributions are as follows.
\begin{itemize}[leftmargin=4mm,topsep=0mm,itemsep=0mm]
    \item We give complete proofs for the OB and TB identification results of \cite{ying2022proximal}; they sketch the argument for the weighting in their Appendix~B.1 without a full proof. These results underlie the DR estimator whose double robustness is established by \cite{ying2022proximal}.
    \item We cast whole-curve OB estimation as a single censored conditional-moment problem with inverse probability of censoring weighted (IPCW) pseudo-outcomes, learned by a projected maximum-moment-restriction objective that couples all horizons through a low-rank time kernel, with a targeted TB stage tied to the fitted outcome bridge.
    \item We derive \modelnamePMMR{} and \modelnameNMMR{}, two complementary OB learners for censored survival data namely, a kernel method with a direct regularized solution and a neural method that relaxes the kernel specification and better accommodates harder misspecification.
    \item We evaluate these methods on three complementary benchmarks: the standard and illustrative structural equation model (SEM) simulation, the challenging proximal semi-synthetic ACTG dataset, and the harder real-world Framingham dataset, in which a separate randomized clinical trial (RCT) provides an external benchmark for the treatment effect. We also add modern causal survival baselines and stress tests with weakened proxies and heavier censoring.
\end{itemize}

\section{Related Work}
PCI identifies causal effects under unmeasured confounding through outcome and treatment bridges \citep{miao2018identifying,tchetgen2020introduction,cui2024semiparametric}, which together yield DR estimators \citep{ghassami2022minimax,cui2024semiparametric}.
These estimators remain consistent when either bridge is correct, a property that becomes especially valuable once the bridges are fit with flexible models.
These ideas trace back to negative-control and proxy-variable arguments in epidemiology \citep{kuroki2014measurement} and have since been combined with modern machine-learning components.
Flexible kernel and neural bridge learners~\citep{Muandet2020-yv,Zhang2020-qr,Mastouri2021-rn,xu2021deep,kompa2022deep,wudoubly2024,meng2025coupling}, largely driven by maximum-moment-restriction objectives, target scalar outcomes and do not handle censoring or survival-curve targets.
For right-censored outcomes, \cite{ying2022proximal} identify treatment-specific survival curves with IPCW \citep{robins2000correcting} and estimate them with low-dimensional parametric bridges, and \cite{hu2025proximal} propose OB, TB, and DR estimators, including nonparametric minimax bridges, for the difference in mean survival times.
We combine existing survival-identification results with flexible proximal moment-restriction learning to estimate the time-indexed OB and corresponding counterfactual survival curves, thus without the need for new identification results. Appendix~\ref{app:related} gives an extended related work discussion.

\section{Proximal Assumptions, OB, TB and DR Estimators}\label{sec:background}
Consider the problem of estimating the causal effect of exposure to treatment $A$ on a potentially censored time-to-event outcome of interest $T$, and let $T_e(a)$ for $a\in\{0,1\}$ be the {\em potential event times}, but acknowledge that only the factual $T_e(a)$ is observed, while the alternative counterfactual $T_e(1-a)$ is not \citep{robins1986new}.
In situations where the outcome $T$ is only partially observed, {\em i.e.}, is {\em right censored}, we have $T=T_c$, otherwise $T$ is fully observed and we have $T=T_e$, {\em i.e.}, $T=\min(T_e,T_c)$, where $T_e$ is the true time to event, with $T_e=T_e(A)$ (consistency). %
Assuming in general that there are pre-exposure observed and unobserved confounders $X$ and $U$, respectively, under the {\em proximal causal inference} framework \citep{tchetgen2020introduction}, it is possible to estimate the causal effect if there are pre-exposure proxies associated with the treatment and outcome, denoted as $Z$ and $W$, respectively.
More formally, we assume the following:

$T_e(a)\indep A|U,X \hspace{2mm} {\rm for} \hspace{2mm} a\in\{0,1\}$ \hfill ({\em latent ignorability})

$(W,T_e(a))\indep (A,Z)|U,X$ \hfill ({\em proxy conditional independence})

$T_e\indep T_c|A,X,W,Z$ \hfill ({\em censoring conditional independence})

$0<p(A=a|U,X)<1$ a.s. for $a\in\{0,1\}$ \hfill ({\em positivity})

$P(T_c>t|A,X,W,Z)>0$ a.s for every horizon $t$ of interest \hfill ({\em censoring positivity})

\begin{figure}[t]
\centering

\begin{tikzpicture}[
    scale=2.0,
    transform shape,
    font=\tiny,
    >={Stealth[length=1.1mm]},
    obs/.style={
        circle,
        draw,
        minimum size=4.2mm,
        inner sep=0pt,
        line width=0.35pt
    },
    lat/.style={
        circle,
        draw,
        dashed,
        fill=gray!30,
        minimum size=4.2mm,
        inner sep=0pt,
        line width=0.35pt
    },
    every edge/.style={draw,->,thin}
]

\node[lat] (U) at (-1.45,0) {$U$};

\node[obs] (W) at (-0.35, 0.90) {$W$};
\node[obs] (X) at (-0.35, 0)    {$X$};
\node[obs] (Z) at (-0.35,-0.90) {$Z$};

\node[obs] (T) at (0.80, 0.65) {$T$};
\node[obs] (A) at (0.80,-0.65) {$A$};

\draw[->] (U) -- (W);
\draw[->] (U) -- (X);
\draw[->] (U) -- (Z);
\draw[->] (U) -- (A);
\draw[->] (U) -- (T);

\draw[->] (X) -- (W);
\draw[->] (X) -- (Z);
\draw[->] (X) -- (A);
\draw[->] (X) -- (T);

\draw[->] (Z) -- (A);
\draw[->] (W) -- (T);
\draw[->] (A) -- (T);

\end{tikzpicture}
\caption{Causal DAG for proximal survival setting. $U$ is latent; $T=\min(T_e,T_c)$ is observed. Missing edges encode proxy independence $(W,T_e(a))\indep(A,Z)\mid U,X$ and censoring independence $T_e\indep T_c\mid A,X,W,Z$.}
\label{fig:dag}
\end{figure}
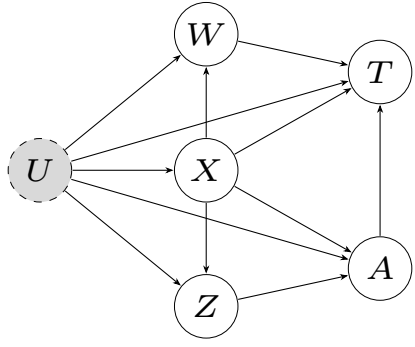

These assumptions, which define the implied directed causal graph in Figure~\ref{fig:dag}, are interpreted as follows:
treatment $A$ and the potential event time $T_e(a)$ are independent given the confounders $(U,X)$;
the treatment and its proxy $Z$ are not directly affected by outcome proxy $W$;
the potential event time $T_e(a)$ is not directly affected by the treatment proxy $Z$;
the event time $T_e$ is not affected by the censoring time $T_c$ given all observed quantities; and
the likelihood of observing either of the exposures $a\in\{0,1\}$ is bounded away from zero.
Moreover, we also assume that proxies $Z$ and $W$ are informative about the unobserved confounder $U$, which is a requirement for the identification guarantees of the outcome and treatment bridge models used to estimate the causal effect \citep{miao2018identifying}.
These models along with formal identification assumption known as {\em completeness} are described later in this section.%

We seek to estimate the causal effect as the difference of marginal counterfactual survival distributions as
\begin{align}\label{eq:psi}
    \begin{split}
        \psi(t) = P(T_e(1)>t)
                - P(T_e(0)>t),
    \end{split}
\end{align}
where the survival distributions will be estimated using the outcome and treatment bridge models described
below.
Note that using hazard (or survival) ratios is also possible but is not considered the main objective due to non-collapsibility issues \citep{daniel2021making}.
This issue makes the comparison of hazard ratios difficult in conditional {\em vs.} marginal models in time-to-event data, so much so that in some cases, it can drastically change the direction and magnitude of estimated causal effect.
For an in-depth discussion on this topic, see \citep{neuhaus1993geometric,sjolander2016note,daniel2021making,gabriel2024propensity}.

\vspace{1em}
\noindent\textbf{Treatment Bridge (TB)}%
\begin{proposition}\label{prp:tb}
    If there exists a TB function $q(Z,X,a)$ that satisfies the following integral equation
    \begin{align}\label{eq:inteq_tb}
        \frac{1}{p(A=a|W,X)} = \mathbb{E}[q(Z,X,a)|W,X,a] \,,
    \end{align}
    almost surely, then
    \begin{align}\label{eq:cdf_cens_tb}
        P(T_e(a)>t) = \mathbb{E}\left[\mathbb{I}(A=a)q(Z,X,a)\frac{\mathbb{I}(T>t)}
            {P(T_c>t|W,Z,X,a)}\right] \,.
    \end{align}
    where $\mathbb{I}(\cdot)$ is the indicator function.
    Moreover, if there were no censoring, then
    \begin{align}\label{eq:cdf_tb}
        \begin{split}
            P(T_e(a)>t)  = \mathbb{E}[\mathbb{I}(A=a)q(Z,X,a)\mathbb{I}(T_e>t)] \,.
        \end{split}
    \end{align}
\end{proposition}
The Proposition~\ref{prp:tb} implies a mechanism for estimating the causal effect in \eqref{eq:psi} without requiring access to the unobserved confounder $U$, but through the treatment bridge $q(Z,X,a)$, which depends only on observed quantities $Z$, $X$ and $A$.
Conceptually, \eqref{eq:cdf_tb} is the proportion of true events that occur after $T_e>t$, weighted by $q(Z,X,a)$, which plays the role of the inverse {\em treatment propensity} based only on observed quantities and restricted to observations consistent with $A=a$.
In practice, we do not have access to $T_e$, but we can use \eqref{eq:cdf_cens_tb}, which is essentially an IPCW version of \eqref{eq:cdf_tb} used to perform statistical adjustment due to censoring \citep{robins2000correcting}.
For example, we can use a Cox proportional hazards model, which is consistent when the censoring model is correctly specified \citep{cox1972regression}.
Consequently, from Proposition~\ref{prp:tb}, $\psi(t)$ is identified by replacing $P(T_e(a)>t)$ with \eqref{eq:cdf_cens_tb} or \eqref{eq:cdf_tb} in \eqref{eq:psi}, and $\hat{\psi}_{\rm TB}(t)$ is its plug-in estimator.
The proof of Proposition~\ref{prp:tb} is presented in Appendix~\ref{app:tb} and requires the assumption about the existence of the treatment bridge in \eqref{eq:inteq_tb}, the assumptions above and completeness {\em i.e.}, for any square-integrable function $g_u(\cdot)$ and for any $A$, $\mathbb{E}[g_u(U)|Z,X,a] = 0$ almost surely if and only if $g_u(U) = 0$ almost surely; and for any square-integrable function $g_z(\cdot)$ and for any $A$, $\mathbb{E}[g_z(Z)|W,X,a] = 0$ almost surely if and only if $g_z(Z)=0$ almost surely. Proposition~\ref{prp:tb} also uses that $\mathbb{E}[g_u(U)|W,X,a]=0$ almost surely only if $g_u(U)=0$ almost surely.
A result similar to Proposition~\ref{prp:tb} was presented in \cite{ying2022proximal}, with a proof sketch in their Appendix~B.1; we present here a complete proof.

Note that it is also possible to account for censoring in \eqref{eq:cdf_tb} using a pseudo-time estimator~\citep{andersen2003generalised}, as in~\citep{wang2018simple} for the case of no unobserved confounders; however, we leave its extension to the proximal causal inference framework as future work.
The next lemma rewrites the treatment-bridge condition as an unconditional balancing relation over functions of \((W,X)\). This form is useful because it connects the abstract bridge equation in Proposition~\ref{prp:tb} to the weighted estimator used in practice.
\begin{lemma}%
\label{lem:tbv2}
If there exists a TB $q(z,a,x)$ such that, for every square-integrable $f(W,X)$,
\begin{equation}
\begin{split}
    &\E\!\left[\mathbb{I}\{A=a\}q(Z,a,X)f(W,X)\right]
    =\E[f(W,X)],
\end{split}
\label{eq:tb-moment}
\end{equation}
\[
    then \quad P(T_e(a)>t) = \mathbb{E}\left[\mathbb{I}(A=a)q(Z,X,a)
    \frac{\mathbb{I}(T>t)}
        {P(T_c>t|W,Z,X,a)}\right].
\]
\end{lemma}

Intuitively, \(q(Z,X,a)\) reweights the treated arm so that functions of \((W,X)\) are balanced as if treatment \(a\) had been assigned independently of latent confounding. This balancing property is what justifies the IPCW-weighted identification formula \eqref{eq:cdf_cens_tb}.
The proof of Lemma~\ref{lem:tbv2} is analogous to the proof of Proposition~\ref{prp:tb} and similar to that in \citep{zhang2025neural}.

\vspace{1em}
\noindent\textbf{Outcome Bridge (OB)}%
\begin{proposition}\label{prp:ob}
    If there exists an OB function $h(t,W,X,a)$ satisfying the following integral equation
    \begin{align}\label{eq:inteq_ob}
        \begin{split}
            P(T_e>t|Z,X,a)
             = \mathbb{E}[h(t,W,X,a)|Z,X,a],
        \end{split}
    \end{align}
    almost surely, then
    \begin{align}
        P(T_e>t|Z,X,a) &= \mathbb{E}\left[\frac{\mathbb{I}(T>t)}
            {P(T_c>t|W,Z,X,a)}\middle|Z,X,a \right], \textit{~and},\label{eq:cdf_cens_ob}\\
        P(T_e(a)>t) &= \mathbb{E}[h(t,W,X,a)].\label{eq:do_ob}
    \end{align}
\end{proposition}
Similarly to Proposition~\ref{prp:tb}, Proposition~\ref{prp:ob} implies a mechanism to estimate the causal effect in \eqref{eq:psi} through the outcome bridge $h(t,W,X,a)$, thus without the need for $U$, {\em i.e.}, the unavailable unobserved confounder.
In practice, assuming access to an estimate $\hat{P}(T_c>t|W,Z,X,a)$, we can optimize the integral equation in \eqref{eq:inteq_ob} using \eqref{eq:cdf_cens_ob} to obtain $\hat{\psi}_{\rm OB}$ by replacing $P(T_e(a)>t)$ with \eqref{eq:do_ob} in \eqref{eq:psi}.
There are several ways to optimize $h(t,W,X,a)$, but here we consider either that \eqref{eq:inteq_ob} is optimized for a single value of $t$ of interest, or in general, over $t>0$.
The proof of Proposition~\ref{prp:ob} is presented in Appendix~\ref{app:ob} and requires the assumption about the existence of the outcome bridge in \eqref{eq:inteq_ob}, the assumptions above and completeness.
\vspace{1em}
\\
\textbf{Targeted Learning of TB using OB}

Lemma~\ref{lemma:obtb} gives a more targeted characterization for learning the treatment bridge once an outcome bridge \(\hat h\) has been fit. It motivates our practical \(q\)-learning stage because it replaces the full balancing family of Lemma~\ref{lem:tbv2} with moments tied directly to \(\hat h_t\).
\begin{lemma}%
\label{lemma:obtb}
For an OB $h(t,w,a,x)$ that solves \eqref{eq:inteq_ob}, if there exists a function $ q(z,a,x)$, for every $a\in\{0,1\}$ and $t\in\cT$, such that
\begin{gather}
\begin{split}
    &\E[h(t,W,a,X)\{1
     -\one\{A=a\} q(Z,a,X)\}]=0,
\end{split}
\label{eq:align-pop}
\end{gather}
    \[
        then\quad P(T_e(a)>t) = \mathbb{E}\left[\mathbb{I}(A=a)q(Z,X,a)
         \frac{\mathbb{I}(T>t)}
            {P(T_c>t|W,Z,X,a)}\right] \,.
    \]
\end{lemma}
Intuitively, this condition aligns the weighted sample with the outcome
bridge at time $t$, yielding a treatment-bridge condition targeted to the
survival estimand.
The proof of Lemma~\ref{lemma:obtb} is given in Appendix~\ref{app:lemma2}.

\vspace{1em}
\noindent\textbf{Doubly-Robust (DR) Estimator}%
\begin{theorem}[\citep{ying2022proximal}]\label{thm:dr}
    If either $q(Z,X,a)$ solves \eqref{eq:inteq_tb} or $h(t,W,X,a)$ solves \eqref{eq:inteq_ob}, then
    \begin{align}\label{eq:do_dr}
        P(T_e(a)>t) = \mathbb{E}\Bigg[ h(t,W,X,a)
         + \mathbb{I}(A=a)q(Z,X,a)
         \Bigg(\frac{\mathbb{I}(T>t)}{P(T_c>t|W,Z,X,a)}
        -h(t,W,X,a) \Bigg)\Bigg] \,,
    \end{align}
    so $\psi(t)$ is identified by replacing $P(T_e(a)>t)$ in \eqref{eq:psi} with \eqref{eq:do_dr}. Under the regularity conditions of \cite{ying2022proximal}, the plug-in estimator $\hat{\psi}_{\rm DR}(t)$ is then consistent for each fixed $t$ if $\hat{P}(T_c>t|W,Z,X,a)$ is consistent and either $\hat q$ or $\hat h$ converges to a solution of \eqref{eq:inteq_tb} or \eqref{eq:inteq_ob}, respectively.
\end{theorem}
The proof of Theorem~\ref{thm:dr} can be found in \cite{ying2022proximal}, which requires the assumptions above, other regularity conditions described there and importantly Propositions \ref{prp:tb} and \ref{prp:ob}, for which \cite{ying2022proximal} sketch the treatment-bridge argument (their Appendix~B.1), and which we prove in full in Appendices \ref{app:tb} and \ref{app:ob}.
Theorem~\ref{thm:dr} is the survival counterpart of proximal double robustness, which implies that the correctness of either bridge suffices once censoring is handled consistently.
In practice, this theorem motivates reporting both OB and DR, while treating TB mainly as a diagnostic of treatment-bridge quality, because TB inherits the full variability of the learned weights, whereas DR uses $q$ only in the augmentation term, where its error is multiplied by the outcome-bridge residual.
Our implementation, therefore, learns $q$ through a bridge-alignment objective
in ~\eqref{eq:align-pop}, together with an additional normalization constraint.
By Lemma~\ref{lemma:obtb}, this is justified when $\hat h$ is consistent for an outcome bridge; otherwise, the learned $q$ need not be a valid treatment bridge, so our implementation need not inherit the full double robustness of Theorem~\ref{thm:dr}.

\section{Model Specification}
\label{sec:learning}

The OB in \eqref{eq:inteq_ob}, combined with \eqref{eq:cdf_cens_ob}, is a conditional moment restriction on observed data, $\E[Y(t)-h(t,W,A,X)\mid A,Z,X]=0$ with the IPCW pseudo-outcome $Y(t)=\mathbb{I}(T>t)/P(T_c>t|A,X,W,Z)$, so $h$ can be learned although $T_e$ is censored.
Both \modelnamePMMR{} and \modelnameNMMR{} solve this restriction through the same population-level objective; they differ only in how they parameterize the outcome bridge $h$.
To convert the conditional restriction into a scalar loss, we follow the maximum-moment-restriction (MMR) framework \cite{Muandet2020-yv,Mastouri2021-rn}. Define the {\em bridge residual} $R_t(h)=Y(t)-h(t,W,A,X)$. For any measurable test function $g(A,Z,X)$, the bridge equation implies $\E[R_t(h)\,g(A,Z,X)]=0$. Because this must hold for {\em all} such $g$, we take the worst case over a reproducing kernel Hilbert space (RKHS) $\cG$ induced by a kernel $k_c:\,(\mathcal{A}\times\mathcal{Z}\times\mathcal{X})^2\to\mathbb{R}$ on the conditioning variables. Similarly, to enforce restriction simultaneously across all time $t\in\cT$, we introduce a time kernel $k_t:\,\cT^2\to\mathbb{R}$ and integrate over a probability measure $\mu(t)$ on $\cT$.

Let $(t,A,X,Z,W)$ and $(t',A',X',Z',W')$ be independent copies, with $(A,X,Z,W),\allowbreak(A',X',Z',W')$ from the observed-data distribution and $t,t'\sim\mu(\cal T)$ independent of the data. The population risk is
$\mathcal{R}(h)=\E[R_t(h)R_{t'}(h)\,k_c((A,Z,X),(A',Z',X'))\,k_t(t,t')]$.

The kernels \(k_c\) and \(k_t\) encode two forms of test-function richness: \(k_c\) probes the conditioning space \((A,Z,X)\), and \(k_t\) couples residual moments across time horizons. This factored structure is also computationally convenient, because it leads to separate Gram matrices for the conditioning variables and the time grid, which both learners exploit.

Given i.i.d.\ samples $\{(\tilde t_i,a_i,x_i,z_i,w_i)\}_{i=1}^n$
from some observed data distribution, below we describe the estimation procedure for outcome and treatment bridges, $h$ and $q$, respectively.
The continuous-time measure $\mu(\cal T)$ in $\mathcal{R}(h)$ is approximated over a finite and pre-defined evaluation grid $\{t_m\}_{m=1}^M$,  and we use $y_i(t_m)=\mathbb{I}\{\tilde t_i>t_m\}/\hat P(T_c>t_m\mid a_i,x_i,w_i,z_i)$ to denote the IPCW pseudo-outcome, where $\hat P(T_c>t\mid a,x,w,z)$ is a Cox proportional hazards model for the censoring time with covariates $(A,X,W,Z)$.%

For the learned bridges $\hat h$ and $\hat q$, our flexible bridge families induce three estimators:
$\hat S_a^{\mathrm{OB}}(t)=\bar h_a(t)$,
$\hat S_a^{\mathrm{TB}}(t)=\frac{1}{n}\sum_{i=1}^n \mathbb{I}\{A_i=a\}\hat q(z_i,a,x_i)y_i(t)$,
and
$\hat S_a^{\mathrm{DR}}(t)=\bar h_a(t)+\frac{1}{n}\sum_{i=1}^n \mathbb{I}\{A_i=a\}\hat q(z_i,a,x_i)\big(y_i(t)-\hat h(t,w_i,a,x_i)\big)$,
where $\bar h_a(t)=\frac{1}{n}\sum_{i=1}^n \hat h(t,w_i,a,x_i)$.
The corresponding causal contrast estimator is
$
\hat\psi^\star(t)=\hat S_1^\star(t)-\hat S_0^\star(t), \star \in \{\mathrm{OB},\mathrm{TB},\mathrm{DR}\}
$.
We reserve labels POR/PIPW/PDR for the semiparametric estimators of \citet{ying2022proximal}. For our flexible bridge families, the names OB/TB/DR emphasize that the weighted estimator is learned through a treatment bridge rather than through a conventional propensity model.
We also report self-normalized H\'ajek~\citep{hajek1971comment} variants, since self-normalization can mitigate instability from extreme weights in weighted estimators, and in our experiments it generally reduces their error (Section~\ref{sec:experiments}). More details are provided in Appendix~\ref{app:hajek}. As a check on survival-curve monotonicity,
Appendix~\ref{app:robustness} shows that monotone rearrangement~\citep{chernozhukov2009improving} does not affect the reported results at the displayed precision.

We can use the kernel-based MMR population risk $\mathcal{R}(h)$ to obtain the following $n$-sample empirical risk
\begin{equation}
\begin{aligned}
&\hat{\mathcal{R}}_{r_t}(h)\;=\;
\frac{1}{n^2\,r_t}
\operatorname{tr}\!\Big(\big({\bf R}_h\,{\bf C}_t\big)^\top\! {\bf K}_c\,\big({\bf R}_h\,{\bf C}_t\big)\Big),
\end{aligned}
\label{eq:common-projected-risk}
\end{equation}
where ${\bf R}_h\in\mathbb{R}^{n\times K}$ is the matrix of outcome bridge residuals with elements $[{\bf R}_h]_{im}=y_i(t_m)-h_{t_m}(w_i,a_i,x_i)$;
${\bf C}_t\in\mathbb{R}^{K\times r_t}$ is the $r_t$-rank Nystr\"om approximation of the full time Gram matrix ${\bf K}_t\in\mathbb{R}^{M\times M}$ evaluated at $\{t_m\}_{m=1}^M$ and a set of $r$ {\em time landmark} points $\{t_\ell^{\rm lm}\}_{\ell=1}^{r_t}$ uniformly spaced over the range of $\{t_m\}_{m=1}^M$, {\em i.e.}, ${\bf C}_t$ has elements $[C_t]_{m,\ell}=k_t(t_m,t_\ell^{\rm lm})$, which form the approximation ${\bf K}_t\approx {\bf C}_t{\bf C}_t^\top$~\citep{williams2000using};
and ${\bf K}_c\in\mathbb{R}^{n\times n}$ is the full conditioning Gram matrix with elements $[K_c]_{ij} = k_c ((a_i,z_i,x_i),\,(a_j,z_j,x_j))$.

Note that the components of $\mathcal{R}(h)$ have been reordered using the properties of the trace $\operatorname{tr}(\cdot)$ operator and that conceptually, the product ${\bf R}_h{\bf C}_t\in\mathbb{R}^{n\times r_t}$ projects the residuals ${\bf R}_h$ onto the $r_t$ time basis functions ${\bf C}_t$, reducing the per-step cost from $O(nM^2)$ to $O(nMr_t)$.
Moreover, the conditioning Gram matrix ${\bf K}_c$ then tests whether the projected residuals are (not) correlated with the conditioning variables $(A,Z,X)$, as required by the outcome bridge identification condition in~\eqref{eq:inteq_ob}.
In practice, we use radial basis function (RBF) kernel functions.
For the conditioning kernel, we use \emph{treatment-matched} kernels, namely observations are compared only within the same treatment arm,
$k_c((a_i,z_i,x_i),(a_j,z_j,x_j))=\mathbb{I}\{a_i=a_j\}\exp(-\|v_i-v_j\|^2/(2\sigma_c^2))$,
where $v_i=(z_i,x_i)$, and $\sigma_c$ is set to the median heuristic on the training set as in \cite{Mastouri2021-rn}.
The time kernel is $k_t(t_m,t_\ell^{\rm lm})=\exp(-(t_m-t_\ell^{\rm lm})^2/(2\sigma_t^2))$, with $\sigma_t$ determined analogously.

The two approaches introduced in the following, namely \modelnamePMMR{} and \modelnameNMMR{} learn the outcome bridge by minimizing a kernel-based MMR approach using the empirical risk objective in~\eqref{eq:common-projected-risk}.
They share the same conditioning kernel $k_c(\cdot,\cdot)$ and time kernel $k_t(\cdot,\cdot)$, with corresponding Gram matrices ${\bf K}_c\in\mathbb{R}^{n\times n}$ and ${\bf K}_t\in\mathbb{R}^{M\times M}$ approximated through ${\bf K}_t\approx {\bf C}_t{\bf C}_t^\top$, respectively.
However, they crucially differ in the manner in which the outcome bridge $h$ is represented and optimized.
In \modelnamePMMR{}, ${\bf K}_c\in\mathbb{R}^{n\times n}$ is instantiated in full whereas in \modelnameNMMR{}, $h_t$ is specified as a neural network and learned using minibatches thus for each minibatch of size $B$ only a Gram sub-matrix ${\bf K}_c\in\mathbb{R}^{B\times B}$ needs to be instantiated. The end of this section compares the two methods.

\begin{table}[t]
\caption{SEM and semi-synthetic ACTG results. Entries report average absolute bias / MAE / Coverage. SEM results vary by sample size; ACTG results use the full ACTG175 cohort ($N{=}2139$) and are averaged over horizons $\{201,401,602,802\}$ days. Appendix Tables \ref{tab:sem-detailed} and \ref{tab:actg-detailed} show the corresponding RMSE values and per-horizon breakdowns. SEM coverage is that of pointwise 95\% Wald intervals with influence-function standard errors over 500 replicates (Cox baselines: replicate-dispersion reconstruction); ACTG coverage is for replicate-dispersion coverage over 30 datasets.}
\label{tab:sem-actg}
\centering
% \small
\resizebox{\textwidth}{!}{
\begin{tabular}{lcccc}
\toprule
 & \multicolumn{3}{c}{\textbf{SEM}} & \textbf{ACTG} \\
\cmidrule(lr){2-4} \cmidrule(lr){5-5}
Method & $N=500$ & $N=1000$ & $N=2000$ & $N=2139$ \\
\midrule
\multicolumn{5}{l}{\emph{Weighted Cox baselines}} \\
Uniform & 0.017 / \textbf{0.028} / 1.00 & 0.016 / 0.022 / 1.00 & 0.018 / 0.020 / 1.00 & 0.072 / 0.072 / 0.34 \\
IPW & 0.030 / 0.035 / 1.00 & 0.029 / 0.031 / 1.00 & 0.030 / 0.031 / 0.67 & 0.068 / 0.071 / 0.43 \\
OW & 0.030 / 0.035 / 1.00 & 0.029 / 0.031 / 1.00 & 0.030 / 0.031 / 0.67 & 0.068 / 0.071 / 0.43 \\
\addlinespace
\multicolumn{5}{l}{\emph{Parametric proximal bridge \citep{ying2022proximal}}} \\
POR & \textbf{0.001} / 0.029 / 0.08 & \textbf{0.000} / \textbf{0.021} / 0.07 & \textbf{0.000} / \textbf{0.016} / 0.07 & 0.156 / 0.184 / 0.38 \\
PIPW & \textbf{0.001} / 0.039 / 1.00 & 0.001 / 0.027 / 1.00 & 0.001 / 0.020 / 0.99 & 0.025 / 0.033 / 0.98 \\
PDR & \textbf{0.001} / 0.039 / \textbf{0.95} & 0.001 / 0.028 / 0.97 & 0.001 / 0.020 / \textbf{0.95} & 0.015 / 0.022 / 0.97 \\
\addlinespace
\multicolumn{5}{l}{\emph{Flexible proximal bridges}} \\
OB-\modelnameNMMR{} & 0.027 / 0.046 / 0.06 & 0.018 / 0.035 / 0.06 & 0.016 / 0.028 / 0.06 & 0.024 / 0.043 / \textbf{0.95} \\
TB-\modelnameNMMR{} & 0.031 / 0.050 / 0.96 & 0.024 / 0.042 / 0.94 & 0.020 / 0.031 / 0.91 & 0.013 / 0.027 / 0.98 \\
DR-\modelnameNMMR{} & 0.019 / 0.038 / 0.94 & 0.014 / 0.028 / 0.94 & 0.012 / 0.022 / 0.91 & 0.006 / 0.016 / 0.99 \\
OB-\modelnamePMMR{} & \textbf{0.001} / 0.037 / 0.44 & 0.002 / 0.026 / 0.41 & 0.001 / 0.020 / 0.36 & 0.010 / 0.020 / 0.94 \\
TB-\modelnamePMMR{} & 0.020 / 0.047 / 0.97 & 0.019 / 0.042 / 0.93 & 0.016 / 0.030 / 0.92 & 0.018 / 0.030 / \textbf{0.95} \\
DR-\modelnamePMMR{} & \textbf{0.001} / 0.039 / \textbf{0.95} & 0.002 / 0.027 / \textbf{0.95} & 0.002 / 0.021 / 0.94 & \textbf{0.002} / \textbf{0.014} / 1.00 \\
\bottomrule
\end{tabular}}
\end{table}

\vspace{1em}
\noindent{\bf \modelnamePMMR{}: Kernel Bridge Representation}
\modelnamePMMR{} generalizes the proximal maximum moment restriction (PMMR) estimator \cite{Mastouri2021-rn} to survival outcomes.
It represents the outcome bridge in a tensor-product RKHS $\cH=\cH_t\otimes\cH_b$, where $\cH_t$ is an RKHS for the time domain $\cT$ with kernel $k_t$ as described above and $\cH_b$ is an RKHS for the bridge input space $(W,A,X)$ with kernel $k_b$.
The latter is a treatment-matched RBF kernel defined analogously to $k_c$ but on $(W,A,X)$ instead of $(A,Z,X)$, which defines the function class in which the outcome bridge resides.
In order to control for the complexity of the outcome bridge we use the following regularized risk,
\begin{equation}
\hat{\mathcal{R}}_{\mathrm{\modelnamePMMR{}}}(h)
\;=\;
\hat{\mathcal{R}}_{r_t}(h)
\;+\;
\lambda\,\|h\|_{\cH}^2,
\label{eq:pmmr-risk}
\end{equation}
where $\hat{\mathcal{R}}_{r_t}(h)$ is the empirical risk in \eqref{eq:common-projected-risk} and $\lambda>0$ is the regularization parameter.

By the representer theorem~\citep{scholkopf2001generalized}, the minimizer of \eqref{eq:pmmr-risk} admits a tensor-product kernel expansion. Resulting $(nM)\times(nM)$ system is reduced using Nystr\"om approximations with $r_s$ sample and $r_t$ time landmarks, yielding a $(r_s r_t)\times(r_s r_t)$ Kronecker-product linear system solved directly by standard linear algebra. Full derivation and prediction formula are given in Appendix~\ref{app:pmmr-details}.

\vspace{1em}
\noindent{\bf \modelnameNMMR{}: Neural Bridge Representation}
\modelnameNMMR{} replaces the RKHS-based bridge representation
with a neural network $h_\theta(t,w,a,x)$ parameterized by $\theta$, generalizing the neural MMR approach of \citet{kompa2022deep} to survival outcomes.
Unlike \modelnamePMMR{}, no bridge kernel $k_b$ is needed since the neural network directly specifies the outcome bridge function.
The \emph{test functions} that enforce the conditional moment restriction, however, remain kernel-based: ${\bf K}_c$ and ${\bf C}_t$ are the same fixed Gram and Nystr\"om matrices defined above for \eqref{eq:common-projected-risk}.

Substituting the neural bridge into the common projected risk~\eqref{eq:common-projected-risk}, the \modelnameNMMR{} learning objective is
\begin{equation}
\begin{aligned}
\hat{\mathcal{R}}_{\mathrm{\modelnameNMMR{}}}(\theta)\;=\;
\frac{1}{n^2\,r_t}
\operatorname{tr}\!\Big(\big({\bf R}_\theta\,{\bf C}_t\big)^\top\! {\bf K}_c\,\big({\bf R}_\theta\,{\bf C}_t\big)\Big),
\end{aligned}
\label{eq:nmmr-projected}
\end{equation}
where ${\bf R}_\theta\in\mathbb{R}^{n\times M}$ collects the residuals $[{\bf R}_\theta]_{im}=Y_i(t_m)-h_\theta(t_m,w_i,a_i,x_i)$.
The objective in \eqref{eq:nmmr-projected} is minimized by minibatch gradient descent, thus within each mini-batch of size $B$, only a $B\times B$ sub-matrix of ${\bf K}_c$ is instantiated, while ${\bf C}_t$ remains fixed.
Finally, the learned $h_\theta$ is used to compute the estimators of OB, TB, and DR analogously to \modelnamePMMR{}.

\vspace{1em}
\noindent{\bf Targeted Treatment-Bridge Learning}
Both learners share the same treatment-bridge stage, fit after $\hat h$ with $\hat h$ held fixed. For each arm $a$, a neural $q_\phi(z,a,x)$ with positive output $1+\exp(\cdot)$ (Table~\ref{tab:architectures}) minimizes the squared empirical alignment residuals of \eqref{eq:align-pop}, summed over the time grid.
We additionally impose the normalization
constraint $\mathbb{E}[\mathbb{I}(A=a)q(Z,a,X)]=1$, which is satisfied
by a valid treatment bridge. This gives:
\begin{equation}
\hat{\mathcal{L}}_a(\phi)={}{\textstyle\sum_{m=1}^{M}}\big(\En[\hat h(t_m,W,a,X)\{1-\mathbb{I}_a\,q_\phi\}]\big)^2
+\lambda_{\mathrm{norm}}\big(\En[\mathbb{I}_a\,q_\phi]-1\big)^2,
\label{eq:q-objective}
\end{equation}
where $\En$ is the empirical mean, $\mathbb{I}_a=\mathbb{I}\{A=a\}$, and $q_\phi=q_\phi(Z,a,X)$. It is minimized by minibatch AdamW, with $\lambda_{\mathrm{norm}}$ fixed or adapted during training (Appendix~\ref{app:implementation}). The $q$ stage is simpler than the $h$ stage by design: $h$ depends on time and three variable blocks and solves an ill-posed integral equation, so its function class matters most, whereas $q$ enters the DR estimator only through the augmentation term.

\vspace{1em}
\noindent{\bf Comparing \modelnamePMMR{} and \modelnameNMMR{}}
Appendix~\ref{app:bridge-comparison} (Table~\ref{tab:bridge-comparison}) contrasts the two learners. Because both minimize the population risk $\mathcal{R}(h)$, their held-out empirical risk \eqref{eq:common-projected-risk}, with common kernels and time grid, can be used to choose between them. On our benchmarks it selects \modelnameNMMR{} on Framingham and \modelnamePMMR{} on ACTG, the more accurate family in each case, and \modelnameNMMR{} on SEM, where the two DR estimators are nearly indistinguishable (MAE 0.022 {\em vs.} 0.021 at $N=2000$). We regard this as practical guidance rather than a theoretical guarantee.

\section{Experiments}
\label{sec:experiments}

We study three benchmarks in increasing difficulty: the SEM simulation of \cite{ying2022proximal}, the proximal semi-synthetic ACTG175 construction, and the real-world Framingham analysis benchmarked against an external randomized trial. We compare against the parametric proximal bridge estimators of \cite{ying2022proximal} and standard weighted Cox baselines (Appendix~\ref{app:robustness} adds four modern causal survival baselines), and evaluate with average absolute bias, MAE, RMSE, and empirical coverage; Appendix~\ref{app:compared-approaches-metrics} gives the full comparison setup and metric definitions. Coverage is dataset-specific (Appendix~\ref{app:metrics}): for SEM, it is the coverage of pointwise 95\% Wald intervals with influence-function standard errors over 500 replicates; for ACTG (30 datasets) and Framingham (30 runs on a fixed cohort), it is replicate-dispersion coverage, i.e., whether the target lies within 1.96 across-replicate standard deviations of each estimate, which reflects bias relative to the Monte Carlo spread rather than interval calibration. Appendix~\ref{app:data} gives SEM/ACTG generation details and Framingham preprocessing pipeline. Training settings for all experiments are given in Appendix~\ref{app:implementation}.

\begin{figure}
\centering
\includegraphics[width=0.68\linewidth]{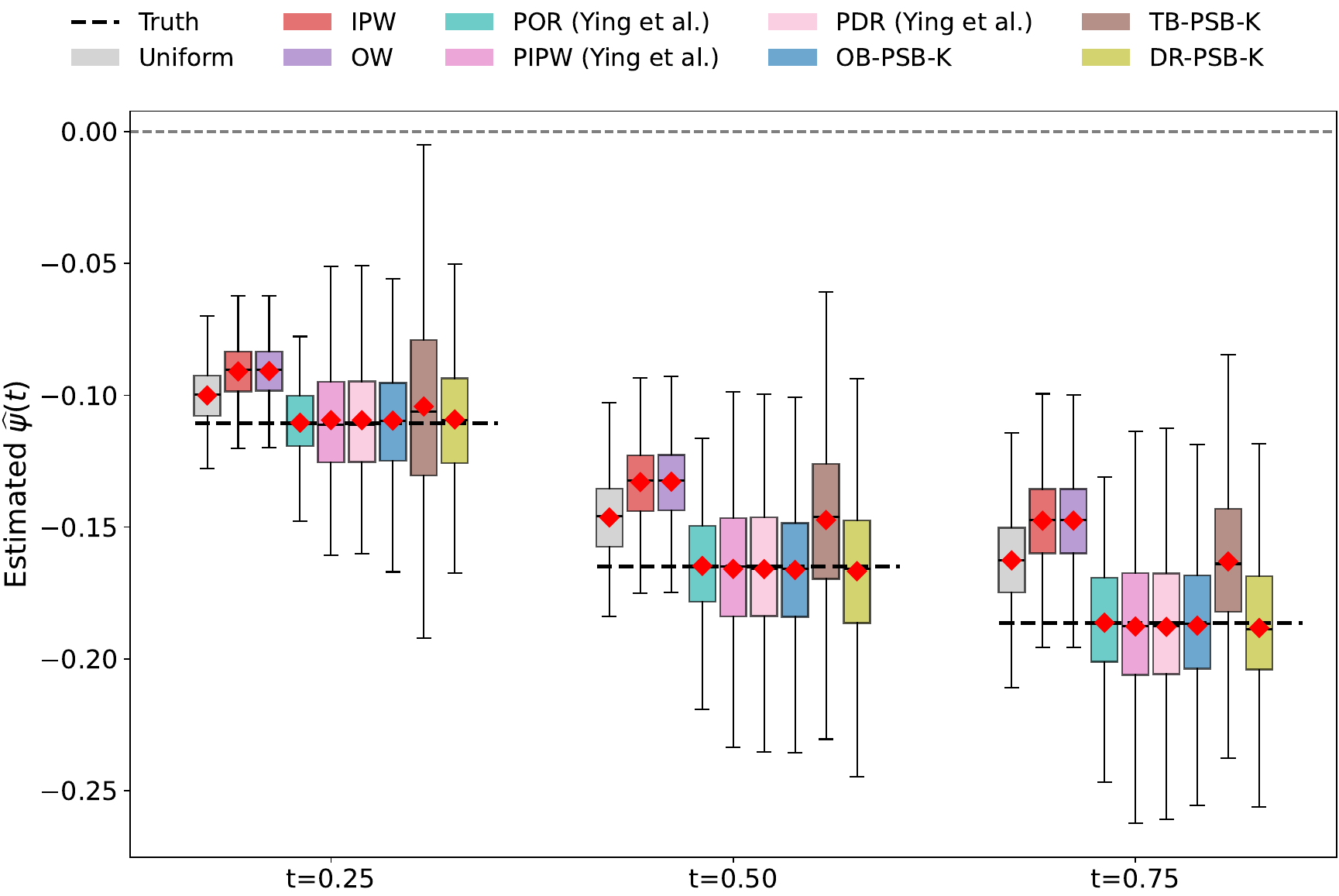}\\
\includegraphics[width=0.68\linewidth]{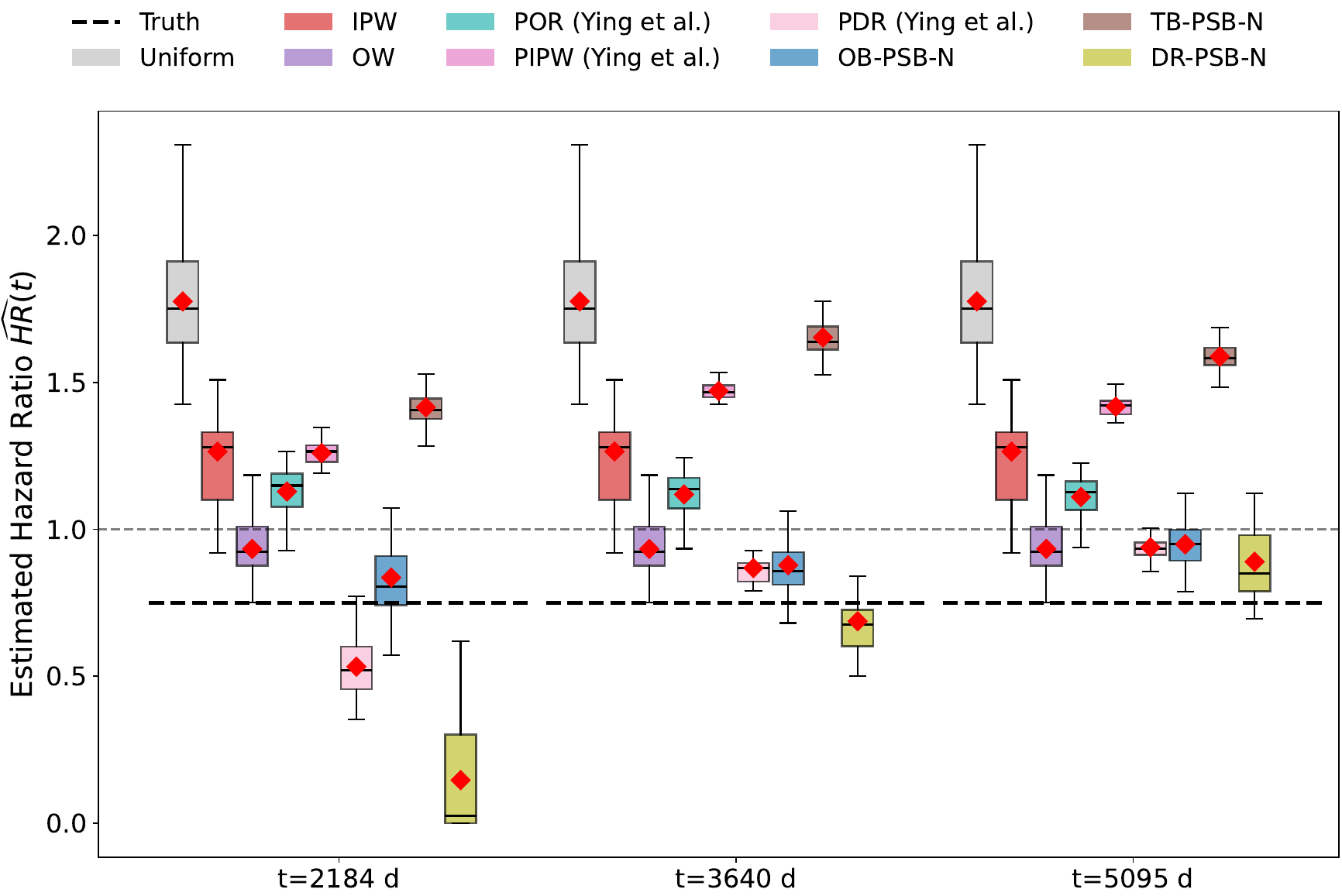}
\caption{Top: SEM at $N=2000$, where \modelnamePMMR{} performs comparably to the parametric bridge in \cite{ying2022proximal}. Bottom: Framingham hazard ratios over 30 runs {\em vs}. the external benchmark $0.75$ (dashed). PIPW, PDR, and TB/DR-\modelnameNMMR{} are shown in their H\'ajek variants (Appendix~\ref{app:hajek}); OB-\modelnameNMMR{} has the smallest average absolute bias.
}
\label{fig:boxplots}
\end{figure}

\vspace{1em}
\noindent\textbf{SEM: matched structure favors the parametric and kernel bridges}
SEM is the easiest regime in the paper: the data is generated from the proximal survival model of \cite{ying2022proximal}, so the identification assumptions hold by construction and reference bridge family is correctly specified.
Table \ref{tab:sem-actg} presents the main results for SEM.
Ying's POR estimator still minimizes average bias and MAE, but it has very poor empirical coverage. By contrast, PDR and DR-\modelnamePMMR{} are close to nominal (0.94--0.97), DR-\modelnameNMMR{} undercovers mildly at $N=2000$ (0.91; Monte Carlo standard error about 0.01), and PIPW is conservative (0.99--1.00). POR and OB undercover severely despite the small bias of POR and OB-\modelnamePMMR{}, which is consistent with influence-function standard errors that treat the fitted bridge as fixed; this first-stage error enters DR only through a product of bridge errors. Among the flexible methods, \modelnamePMMR{} remains the strongest family at every sample size, with DR-\modelnamePMMR{} giving the most balanced accuracy--coverage tradeoff. Appendix Table~\ref{tab:sem-detailed} reports the full Bias/MAE/RMSE/Coverage breakdown, and the top panel of Figure \ref{fig:boxplots} shows the representative $N=2000$ boxplots.

\vspace{1em}
\noindent\textbf{Proximal semi-synthetic ACTG: flexible bridges repair the parametric outcome bridge}
ACTG is the middle-complexity regime. Unlike SEM, it keeps realistic ACTG175 covariates but regenerates the latent severity, proxies, treatment assignment, and hazard surfaces through nonlinear regime switches, so the reference bridge family is no longer well matched to the data-generating mechanism. Table \ref{tab:sem-actg} averages over the four displayed horizons $\{201,401,602,802\}$ days, and Appendix Table~\ref{tab:actg-detailed} provides the per-horizon breakdown. Both flexible bridge learners improve sharply on the reference outcome bridge, reducing the average absolute bias of POR from 0.156 to 0.024 (\modelnameNMMR{}) and 0.010 (\modelnamePMMR{}). The reference treatment bridge fares better, so PDR is protected by double robustness and the gains for the weighted and DR estimators are smaller; after H\'ajek normalization, the reference PIPW and PDR match \modelnameNMMR{} (Appendix Table~\ref{tab:hajek-ablation-full}). \modelnamePMMR{} attains the best average-bias and MAE, while \modelnameNMMR{} remains close; ACTG coverage is replicate-dispersion coverage, so values near 1.00 indicate small bias relative to the spread rather than calibrated intervals. In that sense ACTG sits between the two extremes of the paper: it is harder than SEM, where the matched parametric bridge wins, but not as hostile to kernel structure as Framingham.
This is exactly the regime where flexible bridge learning starts to matter. The appendix generator makes treatment and survival depend on nonlinear proxy transforms, time-varying hazards, and proxy-dependent warping. Those features are difficult for a low-dimensional parametric outcome bridge to absorb, but both \modelnameNMMR{} and \modelnamePMMR{} still track the ground truth well. Without censoring in the same generator, POR's average absolute bias falls from 0.156 to 0.010, between OB-\modelnamePMMR{} (0.004) and OB-\modelnameNMMR{} (0.014), whereas the flexible bridges change by less than 0.01 (Appendix~\ref{app:robustness}). The parametric bridge's failure on ACTG therefore arises mainly under censoring, the setting our survival-specific bridges target.

\vspace{1em}
\noindent\textbf{Framingham: neural bridges are most stable on the observational-to-RCT transfer task}
Framingham is the hardest regime because treatment effects are estimated from observational data, while the benchmark comes from an external randomized trial~\cite{yusuf2016cholesterol} in a different population. Our estimators recover the marginal survival curves $(\hat S_0,\hat S_1)$, so in principle they could be evaluated through differences or ratios; here we obtain hazard-ratios from those marginals as in Appendix \ref{app:metrics}, because the external benchmark is a constant hazard ratio of $0.75$; differences from it reflect both estimation error and differences between the trial and cohort populations.
The bottom panel of Figure \ref{fig:boxplots} shows the resulting distributions, and  Table \ref{tab:framingham-detailed} reports Bias, MAE, RMSE, and replicate-dispersion coverage at 2184, 3640, and 5095 days. \modelnameNMMR{} is the only flexible bridge family that remains stable throughout: OB-\modelnameNMMR{} is the strongest standard-run method overall (Avg.\ abs.\ bias $0.138$, MAE $0.151$), while OW is the best non-proximal baseline and still trails the neural bridge. By contrast, the kernel bridge deteriorates sharply on this challenging real-data task, likely because its single median-heuristic bandwidth cannot adapt to the high-dimensional, mixed-type proxy blocks. Because the 30 runs share one cohort and differ only in the train/validation split and stochastic fitting, their spread reflects computational variability rather than sampling uncertainty, so the coverage values ({\em e.g.}, 0.70 for DR-\modelnameNMMR{}) are not coverage of 95\% confidence intervals.

\vspace{1em}
\noindent\textbf{Modern baselines and stress tests}
Four modern causal survival estimators that assume no unmeasured confounding (survival TMLE, IPCW-AIPW, causal survival forests, and DeepSurv with IPW) have average absolute bias 0.019--0.031 on SEM ($N=2000$) and 0.006--0.086 on ACTG, versus 0.0017 and 0.0021 for DR-\modelnamePMMR{}. As the SEM proxies are weakened, DR-\modelnamePMMR{} is stable at first (bias 0.0021 and coverage 0.93 when the proxy--confounder correlation is 75\% of its baseline) but fails when they are weak (0.0471 and 0.57 at 10\%, worse than the unadjusted Kaplan--Meier contrast); an invalid treatment proxy raises its bias to 0.0128 (coverage 0.90). On ACTG, censoring of up to 60\% and simpler censoring models keep the DR bias between 0.002 and 0.007. Appendix~\ref{app:robustness} (Table~\ref{tab:modern-baselines}) gives the designs and full results (30 replicates each).

\vspace{1em}
\noindent\textbf{H\'ajek normalization}
Appendix Table~\ref{tab:hajek-ablation-full} asks a narrower question: once the bridge family is fixed, does H\'ajek self-normalization help the weighted estimators?  It does. On ACTG it lowers the average absolute bias of all four weighted estimators to 0.004--0.008, most for the parametric PIPW and PDR of \citet{ying2022proximal} (from 0.025 and 0.015) and least for DR-\modelnameNMMR{} (from 0.006). In Framingham, where the hazard ratios can have heavy right tails, it lowers the error of PIPW, PDR, and TB-\modelnameNMMR{}, most visibly for the reference PIPW estimator. There, normalization reduces the RMSE of PIPW from 22.0 to 0.634, but its bias remains 0.633 and its replicate-dispersion coverage is 0.00. Even there, it does not turn weighted estimators into the best overall option, and for \modelnameNMMR{} DR it slightly increases the error (average absolute bias 0.241 to 0.269).  The main tables report the standard forms, following \citet{ying2022proximal}. The effective sample size (ESS) of learned weights \citep{kish1965survey} is 96.8\% of $n$ on ACTG and 95.6--99.1\% on SEM but 51.8\% on Framingham.  We recommend the H\'ajek variants of weighted estimators, particularly when the ESS falls below about half of $n$, as on Framingham. More results are provided in Appendix Table~\ref{tab:hajek-ablation-full}.

\section{Conclusion}
\label{sec:conclusion}
We introduced \modelfullform{} (\modelname{}), a doubly robust proximal framework for right-censored survival outcomes that combines survival-specific proximal identification with flexible kernel and neural bridge learning. The framework identifies treatment-specific survival curves through time-indexed outcome and treatment bridges, and yields OB, TB, and DR estimators built on the doubly robust identification of \cite{ying2022proximal}.
Across three benchmarks of increasing difficulty, the results show that the preferred bridge family depends on the complexity of the underlying confounding structure. \modelnamePMMR{} performs well in the smoother SEM regime and remains competitive on ACTG, while \modelnameNMMR{} is the most stable on the more challenging Framingham task. Overall, these results show that flexible survival-specific bridge learning can improve proximal causal estimation beyond parametric survival bridges.
We discuss the limitations of our framework in Appendix~\ref{app:limitations}.

\subsection*{AI use statement}
Generative AI tools were used to provide feedback on research methodology, assist with code development, and support the drafting and editing of the manuscript. They were also used for brainstorming, literature search, and improving the clarity and presentation of the paper. All AI-assisted content, mathematical arguments, code, and scientific claims were reviewed and verified by the authors. The authors take full responsibility for the final content of this work.

\bibliographystyle{plainnat}
\bibliography{references}

@inproceedings{Mastouri2021-rn,
  title={Proximal Causal Learning with Kernels: Two-Stage Estimation and Moment Restriction},
  author={Mastouri, Afsaneh and Zhu, Yuchen and Gultchin, Limor and Korba, Anna and Silva, Ricardo and Kusner, Matt and Gretton, Arthur and Muandet, Krikamol},
  booktitle={Proceedings of the 38th International Conference on Machine Learning},
  series={Proceedings of Machine Learning Research},
  volume={139},
  pages={7512--7523},
  year={2021}
}

@article{Zhang2020-qr,
  title={Maximum Moment Restriction for Instrumental Variable Regression},
  author={Zhang, Rui and Imaizumi, Masaaki and Sch{\"o}lkopf, Bernhard and Muandet, Krikamol},
  journal={arXiv preprint arXiv:2010.07684},
  year={2020}
}

@inproceedings{Muandet2020-yv,
  title={Kernel Conditional Moment Test via Maximum Moment Restriction},
  author={Muandet, Krikamol and Jitkrittum, Wittawat and K{\"u}bler, Jonas},
  booktitle={Proceedings of the 36th Conference on Uncertainty in Artificial Intelligence},
  series={Proceedings of Machine Learning Research},
  volume={124},
  pages={41--50},
  year={2020}
}

@article{li2018balancing,
  title={Balancing covariates via propensity score weighting},
  author={Li, Fan and Morgan, Kari Lock and Zaslavsky, Alan M},
  journal={Journal of the American Statistical Association},
  volume={113},
  number={521},
  pages={390--400},
  year={2018},
  publisher={Taylor \& Francis}
}

@article{cao2009improving,
  title={Improving efficiency and robustness of the doubly robust estimator for a population mean with incomplete data},
  author={Cao, Weihua and Tsiatis, Anastasios A and Davidian, Marie},
  journal={Biometrika},
  volume={96},
  number={3},
  pages={723--734},
  year={2009},
  publisher={Oxford University Press}
}

@article{rosenbaum1983central,
  title={The central role of the propensity score in observational studies for causal effects},
  author={Rosenbaum, Paul R and Rubin, Donald B},
  journal={Biometrika},
  volume={70},
  number={1},
  pages={41--55},
  year={1983},
  publisher={Oxford University Press}
}

@article{hammer1996trial,
  title={A trial comparing nucleoside monotherapy with combination therapy in HIV-infected adults with CD4 cell counts from 200 to 500 per cubic millimeter},
  author={Hammer, Scott M and Katzenstein, David A and Hughes, Michael D and Gundacker, Holly and Schooley, Robert T and Haubrich, Richard H and Henry, W Keith and Lederman, Michael M and Phair, John P and Niu, Manette and others},
  journal={New England Journal of Medicine},
  volume={335},
  number={15},
  pages={1081--1090},
  year={1996},
  publisher={Mass Medical Soc}
}

@article{williams2000using,
  title={Using the Nystr{\"o}m method to speed up kernel machines},
  author={Williams, Christopher and Seeger, Matthias},
  journal={Advances in neural information processing systems},
  volume={13},
  year={2000}
}

@article{buchanan2014worth,
  title={Worth the weight: using inverse probability weighted Cox models in AIDS research},
  author={Buchanan, Ashley L and Hudgens, Michael G and Cole, Stephen R and Lau, Bryan and Adimora, Adaora A and Women's Interagency HIV Study},
  journal={AIDS research and human retroviruses},
  volume={30},
  number={12},
  pages={1170--1177},
  year={2014},
  publisher={Mary Ann Liebert, Inc. 140 Huguenot Street, 3rd Floor New Rochelle, NY 10801 USA}
}

@article{schemper2009estimation,
  title={The estimation of average hazard ratios by weighted Cox regression},
  author={Schemper, Michael and Wakounig, Samo and Heinze, Georg},
  journal={Statistics in Medicine},
  volume={28},
  number={19},
  pages={2473--2489},
  year={2009},
  publisher={Wiley Online Library}
}

@article{hajek1971comment,
  title={Comment on “an essay on the logical foundations of survey sampling, part one”},
  author={H{\'a}jek, Jaroslav},
  journal={The foundations of survey sampling},
  volume={236},
  year={1971},
  publisher={Holt, Rinehart, and Winston Toronto}
}

@inproceedings{scholkopf2001generalized,
  title={A generalized representer theorem},
  author={Sch{\"o}lkopf, Bernhard and Herbrich, Ralf and Smola, Alex J},
  booktitle={International conference on computational learning theory},
  pages={416--426},
  year={2001},
  organization={Springer}
}

@article{zhang2025neural,
  title={Neural estimation of treatment bridge functions for proximal causal inference},
  author={Zhang, Bingxi and Shen, Tao and Cui, Yifan},
  journal={Statistical Analysis and Data Mining: An ASA Data Science Journal},
  volume={18},
  number={5},
  pages={e70045},
  year={2025},
  publisher={Wiley Online Library}
}

@book{imbens2015causal,
  title={Causal inference in statistics, social, and biomedical sciences},
  author={Imbens, Guido W and Rubin, Donald B},
  year={2015},
  publisher={Cambridge university press}
}

@misc{hernan2010causal,
  title={Causal inference},
  author={Hern{\'a}n, Miguel A and Robins, James M},
  year={2010},
  publisher={CRC Boca Raton, FL}
}

@book{kalbfleisch2002statistical,
  title={The statistical analysis of failure time data},
  author={Kalbfleisch, John D and Prentice, Ross L},
  year={2002},
  publisher={John Wiley \& Sons}
}

@incollection{rosenbaum2002observational,
  title={Observational studies},
  author={Rosenbaum, Paul R},
  booktitle={Observational studies},
  pages={1--17},
  year={2002},
  publisher={Springer}
}

@article{denz2023comparison,
  title={A comparison of different methods to adjust survival curves for confounders},
  author={Denz, Robin and Klaa{\ss}en-Mielke, Renate and Timmesfeld, Nina},
  journal={Statistics in medicine},
  volume={42},
  number={10},
  pages={1461--1479},
  year={2023},
  publisher={Wiley Online Library}
}

@book{pearl2016causal,
  title={Causal inference in statistics: A primer},
  author={Pearl, Judea and Glymour, Madelyn and Jewell, Nicholas P},
  year={2016},
  publisher={John Wiley \& Sons}
}

@article{rosenbaum1984association,
  title={From association to causation in observational studies: The role of tests of strongly ignorable treatment assignment},
  author={Rosenbaum, Paul R},
  journal={Journal of the American Statistical Association},
  volume={79},
  number={385},
  pages={41--48},
  year={1984},
  publisher={Taylor \& Francis}
}

@article{kuroki2014measurement,
  title={Measurement bias and effect restoration in causal inference},
  author={Kuroki, Manabu and Pearl, Judea},
  journal={Biometrika},
  volume={101},
  number={2},
  pages={423--437},
  year={2014},
  publisher={Oxford University Press}
}

@article{miao2018identifying,
  title={Identifying causal effects with proxy variables of an unmeasured confounder},
  author={Miao, Wang and Geng, Zhi and Tchetgen Tchetgen, Eric J.},
  journal={Biometrika},
  volume={105},
  number={4},
  pages={987--993},
  year={2018}
}

@article{robins1986new,
  title={A new approach to causal inference in mortality studies with a sustained exposure period—application to control of the healthy worker survivor effect},
  author={Robins, James},
  journal={Mathematical modelling},
  volume={7},
  number={9-12},
  pages={1393--1512},
  year={1986},
  publisher={Elsevier}
}

@article{tchetgen2020introduction,
  title={An introduction to proximal causal learning},
  author={Tchetgen, Eric J Tchetgen and Ying, Andrew and Cui, Yifan and Shi, Xu and Miao, Wang},
  journal={arXiv preprint arXiv:2009.10982},
  year={2020}
}

@article{ying2022proximal,
  title={Proximal causal inference for marginal counterfactual survival curves},
  author={Ying, Andrew and Cui, Yifan and Tchetgen, Eric J Tchetgen},
  journal={arXiv preprint arXiv:2204.13144},
  year={2022}
}

@article{cox1972regression,
  title={Regression models and life-tables},
  author={Cox, David R},
  journal={Journal of the Royal Statistical Society: Series B (Methodological)},
  volume={34},
  number={2},
  pages={187--202},
  year={1972},
  publisher={Wiley Online Library}
}

@article{robins2000correcting,
  title={Correcting for noncompliance and dependent censoring in an AIDS clinical trial with inverse probability of censoring weighted (IPCW) log-rank tests},
  author={Robins, James M and Finkelstein, Dianne M},
  journal={Biometrics},
  volume={56},
  number={3},
  pages={779--788},
  year={2000},
  publisher={Wiley Online Library}
}

@article{daniel2021making,
  title={Making apples from oranges: Comparing noncollapsible effect estimators and their standard errors after adjustment for different covariate sets},
  author={Daniel, Rhian and Zhang, Jingjing and Farewell, Daniel},
  journal={Biometrical Journal},
  volume={63},
  number={3},
  pages={528--557},
  year={2021},
  publisher={Wiley Online Library}
}

@article{sjolander2016note,
  title={A note on the noncollapsibility of rate differences and rate ratios},
  author={Sj{\"o}lander, Arvid and Dahlqwist, Elisabeth and Zetterqvist, Johan},
  journal={Epidemiology},
  volume={27},
  number={3},
  pages={356--359},
  year={2016},
  publisher={LWW}
}

@article{neuhaus1993geometric,
  title={A geometric approach to assess bias due to omitted covariates in generalized linear models},
  author={Neuhaus, John M and Jewell, Nicholas P},
  journal={Biometrika},
  volume={80},
  number={4},
  pages={807--815},
  year={1993},
  publisher={Oxford University Press}
}

@article{gabriel2024propensity,
  title={Propensity weighting plus adjustment in proportional hazards model is not doubly robust},
  author={Gabriel, Erin E and Sachs, Michael C and Waernbaum, Ingeborg and Goetghebeur, Els and Blanche, Paul F and Vansteelandt, Stijn and Sj{\"o}lander, Arvid and Scheike, Thomas},
  journal={Biometrics},
  volume={80},
  number={3},
  pages={ujae069},
  year={2024},
  publisher={Oxford University Press}
}

@article{cui2024semiparametric,
  title={Semiparametric proximal causal inference},
  author={Cui, Yifan and Pu, Hongming and Shi, Xu and Miao, Wang and Tchetgen Tchetgen, Eric},
  journal={Journal of the American Statistical Association},
  volume={119},
  number={546},
  pages={1348--1359},
  year={2024},
  publisher={Taylor \& Francis}
}

@article{xu2021deep,
  title={Deep proxy causal learning and its application to confounded bandit policy evaluation},
  author={Xu, Liyuan and Kanagawa, Heishiro and Gretton, Arthur},
  journal={Advances in Neural Information Processing Systems},
  volume={34},
  pages={26264--26275},
  year={2021}
}

@inproceedings{meng2025coupling,
  title={Coupling Generative Modeling and an Autoencoder with the Causal Bridge},
  author={Meng, Ruolin and Chung, Ming-Yu and Brahma, Dhanajit and Henao, Ricardo and Carin, Lawrence},
  booktitle={The Thirty-ninth Annual Conference on Neural Information Processing Systems},
  year={2025}
}

@article{kompa2022deep,
  title={Deep learning methods for proximal inference via maximum moment restriction},
  author={Kompa, Benjamin and Bellamy, David and Kolokotrones, Tom and Beam, Andrew and others},
  journal={Advances in Neural Information Processing Systems},
  volume={35},
  pages={11189--11201},
  year={2022}
}

@inproceedings{wudoubly2024,
  title={Doubly Robust Proximal Causal Learning for Continuous Treatments},
  author={Wu, Yong and Fu, Yanwei and Wang, Shouyan and Sun, Xinwei},
  booktitle={The Twelfth International Conference on Learning Representations},
  year={2024}
}

@inproceedings{ghassami2022minimax,
  title={Minimax kernel machine learning for a class of doubly robust functionals with application to proximal causal inference},
  author={Ghassami, AmirEmad and Ying, Andrew and Shpitser, Ilya and Tchetgen, Eric Tchetgen},
  booktitle={International conference on artificial intelligence and statistics},
  pages={7210--7239},
  year={2022},
  organization={PMLR}
}

@article{benjamin1994independent,
  title={Independent risk factors for atrial fibrillation in a population-based cohort: the Framingham Heart Study},
  author={Benjamin, Emelia J and Levy, Daniel and Vaziri, Sonya M and D'Agostino, Ralph B and Belanger, Albert J and Wolf, Philip A},
  journal={Jama},
  volume={271},
  number={11},
  pages={840--844},
  year={1994},
  publisher={American Medical Association}
}

@article{yusuf2016cholesterol,
  title={Cholesterol lowering in intermediate-risk persons without cardiovascular disease},
  author={Yusuf, Salim and Bosch, Jackie and Dagenais, Gilles and Zhu, Jun and Xavier, Denis and Liu, Lisheng and Pais, Prem and L{\'o}pez-Jaramillo, Patricio and Leiter, Lawrence A and Dans, Antonio and others},
  journal={New England Journal of Medicine},
  volume={374},
  number={21},
  pages={2021--2031},
  year={2016},
  publisher={Mass Medical Soc}
}

@article{wang2018simple,
  title={A simple, doubly robust, efficient estimator for survival functions using pseudo observations},
  author={Wang, Jixian},
  journal={Pharmaceutical Statistics},
  volume={17},
  number={1},
  pages={38--48},
  year={2018},
  publisher={Wiley Online Library}
}

@article{andersen2003generalised,
  title={Generalised linear models for correlated pseudo-observations, with applications to multi-state models},
  author={Andersen, Per Kragh and Klein, John P and Rosth{\o}j, Susanne},
  journal={Biometrika},
  volume={90},
  number={1},
  pages={15--27},
  year={2003},
  publisher={Oxford University Press}
}

@article{hu2025proximal,
  title={{Proximal Causal Inference for Censored Data with an Application to Right Heart Catheterization Data}},
  author={Hu, Yue and Gao, Yuanshan and Qi, Minhao},
  journal={Stats},
  volume={8},
  number={3},
  pages={66},
  year={2025},
  publisher={MDPI}
}

@article{chernozhukov2009improving,
  title={{Improving point and interval estimators of monotone functions by rearrangement}},
  author={Chernozhukov, Victor and Fern{\'a}ndez-Val, Iv{\'a}n and Galichon, Alfred},
  journal={Biometrika},
  volume={96},
  number={3},
  pages={559--575},
  year={2009}
}

@incollection{rotnitzky2005inverse,
author = {Rotnitzky, Andrea and Robins, James M.},
publisher = {John Wiley \& Sons},
title = {Inverse Probability Weighting in Survival Analysis},
booktitle = {Wiley StatsRef: Statistics Reference Online},
year = {2014},
}

@book{kish1965survey,
  title={{Survey Sampling}},
  author={Kish, Leslie},
  publisher={John Wiley \& Sons},
  address={New York},
  year={1965}
}

@article{cai2020one,
  title={{One-step targeted maximum likelihood estimation for time-to-event outcomes}},
  author={Cai, Weixin and van der Laan, Mark J.},
  journal={Biometrics},
  volume={76},
  number={3},
  pages={722--733},
  year={2020}
}

@article{bai2013doubly,
  title={{Doubly-robust estimators of treatment-specific survival distributions in observational studies with stratified sampling}},
  author={Bai, Xiaofei and Tsiatis, Anastasios A. and O'Brien, Sean M.},
  journal={Biometrics},
  volume={69},
  number={4},
  pages={830--839},
  year={2013},
  publisher={Oxford University Press}
}

@article{cui2023estimating,
  title={{Estimating heterogeneous treatment effects with right-censored data via causal survival forests}},
  author={Cui, Yifan and Kosorok, Michael R. and Sverdrup, Erik and Wager, Stefan and Zhu, Ruoqing},
  journal={Journal of the Royal Statistical Society Series B: Statistical Methodology},
  volume={85},
  number={2},
  pages={179--211},
  year={2023},
  publisher={Oxford University Press US}
}

@article{katzman2018deepsurv,
  title={{DeepSurv: personalized treatment recommender system using a Cox proportional hazards deep neural network}},
  author={Katzman, Jared L. and Shaham, Uri and Cloninger, Alexander and Bates, Jonathan and Jiang, Tingting and Kluger, Yuval},
  journal={BMC Medical Research Methodology},
  volume={18},
  number={1},
  pages={24},
  year={2018},
  publisher={Springer}
}

\clearpage

\startcontents[appendices]
\printcontents[appendices]{l}{1}{\section*{Appendix Contents}}

\newpage
\appendix

\section{Limitations} \label{app:limitations}
The guarantees rely on strong proximal assumptions, especially completeness, which are not directly testable from observed data. The availability and quality of suitable proxies \(W\) and \(Z\) are application-dependent, and the Framingham evaluation is limited to hazard-ratio summaries rather than full survival-curve contrasts. In addition, the kernel bridge is less reliable on harder real-world tasks, motivating future work on sensitivity analyses for completeness violations and adaptive hybrids between kernel and neural bridges. The theory, like other state-of-the-art works, is at the population level: we do not establish rates, uniform-in-time consistency, or asymptotic distributions for the learned bridges.
Simple IPCW requires a correct censoring model and conditionally independent censoring; augmented IPCW \citep{rotnitzky2005inverse} would improve efficiency and robustness by changing only the pseudo-outcome, whereas latent censoring-outcome dependence would require additional structure. Moreover, the diagnostics can flag weak proxies or unstable weights but cannot certify proxy validity or completeness.

\section{Extended Related Work}\label{app:related}
The literature on PCI builds on foundational work in causal identification with proxy variables, primarily focusing on addressing the breakdown of the no unmeasured confounding assumption in observational studies.
Early PCI methods \citep{miao2018identifying,tchetgen2020introduction} established the nonparametric identification of the average treatment effect (ATE) directly modeling the (continuous or binary) outcome using an {\em outcome bridge} function.
Subsequent work expanded PCI to complex designs, including generalized regression approaches and specifications based on neural networks \citep{xu2021deep,ghassami2022minimax,kompa2022deep,wudoubly2024,meng2025coupling}.

More recently, the PCI framework was extended to model the propensity to treatment using a proximal inverse probability-weighted (PIPW) estimator, or {\em treatment bridge}, directly adapted standard IPW techniques to the proximal setting \citep{cui2024semiparametric}.
Furthermore, having access to both the outcome and treatment bridges allowed the introduction of DR estimators \citep{ghassami2022minimax,cui2024semiparametric}, as a means to enhance estimation robustness.
Such a DR estimator guarantees consistency if either the treatment or outcome bridge function is correctly specified.
This framework was then extended to survival outcomes using IPCW \citep{robins2000correcting}, to appropriately handle right-censored data.
However, in \citep{ying2022proximal}:
$i$) complete proofs of the outcome- and treatment-bridge identification results were not given (their Appendix~B.1 sketches the argument for the weighting route, and the DR property is proved in full);
$ii$) their experiments were largely performed on synthetic SEM data for which the outcome and treatment bridge were assumed to match the true underlying generative process; and
$iii$) their results on real-world data lacked an external benchmark to compare their findings.
This work directly addresses these gaps as described in Section~\ref{sec:intro}.

The estimators in \citep{ying2022proximal} rely on low-dimensional parametric bridge forms; however, we generalize the bridge family with flexible kernel and neural learners that are not tied to that parametric specification. Flexible bridge learning has been developed for scalar-outcome proximal problems using kernel and neural moment-restriction ideas \citep{Mastouri2021-rn,Zhang2020-qr,Muandet2020-yv,xu2021deep,kompa2022deep}. Those methods motivate our learning objectives, but they do not directly handle censoring, horizon-specific pseudo-outcomes, or survival-curve targets.
The present work combines these two threads: survival identification from \cite{ying2022proximal} and flexible bridge learning from proximal moment-restriction methods. \cite{hu2025proximal} propose OB, TB, and DR proximal estimators for the difference in mean potential survival times under right censoring, and also formulate nonparametric minimax bridge estimators.
We combine existing survival-identification results with flexible proximal moment-restriction learning to estimate the time-indexed OB and corresponding counterfactual survival curves, thus without the need for new identification results.

\section{Proof of Proposition~\ref{prp:tb}}\label{app:tb}
We first show \eqref{eq:cdf_tb}
\begin{align}
    P(T_e(a)>t) &=
    \int dX \int dU  \ p(T_e(a)>t|U,X) p(U,X) \frac{p(U,X|a)}{p(U,X|a)} \notag \\
    & = \int dX \int dU  \ p(T_e(a)>t|U,X) \frac{p(U,X|a)p(a)}{p(A=a|U,X)} \label{eq:do_tb_1} \\
    % \displaybreak
    & = \int dX \int dU \ p(T_e(a)>t|U,X) \int dZ \ q(Z,X,a) p(Z|U,X,a) p(U,X|a) p(a) \label{eq:do_tb_2} \\
    & = \int dX \int dU \ p(T_e>t|U,X,a) \int dZ \ q(Z,X,a) p(Z|U,X,a) p(U,X|a) p(a) \label{eq:do_tb_3} \\
    & = \int dX \int dU \int dT_e \ \mathbb{I}(T_e>t) \int dZ \ q(Z,X,a) p(T_e|U,X,Z,a) p(Z|U,X,a) p(U,X|a) p(a) \label{eq:do_tb_4} \\
    & = \int dX \int dU \int dT_e \ \mathbb{I}(T_e>t) \int dZ \ q(Z,X,a) p(T_e,U,X,Z|a) p(a) \label{eq:do_tb_5} \\
    & =\int dX \int dT_e \int dZ \ \mathbb{I}(T_e>t) q(Z,X,a) p(T_e,X,Z|a)p(a) \notag \\
    & = \mathbb{E}[\mathbb{I}(A=a)q(Z,X,a)\mathbb{I}(T_e>t)] \,. \label{eq:do_tb_6}
\end{align}
where
\begin{itemize}[leftmargin=4mm,topsep=0mm,itemsep=0mm]
    \item In \eqref{eq:do_tb_1} we have used
    \begin{align*}
        p(U,X|a) = \frac{p(a|U,X)p(U,X)}{p(a)} \,.
    \end{align*}
    \item In \eqref{eq:do_tb_2} we have used
    \begin{align}\label{eq:paux}
        \frac{1}{p(A=a|U,X)} = \int dZ \ q(Z,X,a) p(Z|U,X,a) \,.
    \end{align}
    To show \eqref{eq:paux}
    \begin{align}
        \int dU \ \frac{1}{p(A=a|U,X)}p(U|W,X,a) & = \int dU \ \frac{1}{p(A=a|U,W,X)}p(U|W,X,a) \label{eq:paux_proof_1} \\
        & = \frac{1}{p(A=a|W,X)} \notag \\
        & = \mathbb{E}[q(Z,X,a)|W,X,a] \label{eq:paux_proof_2} \\
        & = \int dU \int dZ \ q(Z,X,a) p(Z|U,X,W,a) p(U|W,X,a) \notag \\
        & = \int dU \int dZ \ q(Z,X,a) p(Z|U,X,a) p(U|W,X,a) \notag \\
        & = \int dU \ \frac{1}{p(A=a|U,X)}p(U|W,X,a) \label{eq:paux_proof_3} \,,
    \end{align}
    where in \eqref{eq:paux_proof_1} we have used $(W,T_e(a))\indep (A,Z)|U,X$, in \eqref{eq:paux_proof_2} we have used
    \begin{align*}
        \frac{1}{p(A=a|W,X)} = \mathbb{E}[q(Z,X,a)|W,X,a] \,,
    \end{align*}
    and in \eqref{eq:paux_proof_3} we have used that by the completeness assumption \citep{cui2024semiparametric}, namely that $\mathbb{E}[g_u(U)|W,X,a]=0$ almost surely only if $g_u(U)=0$ almost surely,
    \begin{align*}
        \frac{1}{p(A=a|U,X)} = \int dZ \ q(Z,X,a) p(Z|U,X,a) \,.
    \end{align*}
    \item In \eqref{eq:do_tb_3} we have used
    \begin{align}\label{eq:do_eq}
        P(T_e(a)>t|U,X) = P(T_e>t|U,X,a) \,,
    \end{align}
    provided that $T_e(a)\indep A|(U,X)$ and $T_e=T_e(A)$, as shown in \citep{robins1986new}.
    \item In \eqref{eq:do_tb_4} we have used
    \begin{align*}
        P(T_e>t|U,X,a) & = \int dT_e \ \mathbb{I}(T_e>t) p(T_e|U,X,a) \\
        & = \int dT_e \ \mathbb{I}(T_e>t) p(T_e|U,X,Z,a) \,,
    \end{align*}
    provided that $(W,T_e(a))\indep (A,Z)|U,X$.
    \item In \eqref{eq:do_tb_5} we have used
    \begin{align*}
        p(T_e,U,X,Z|a) = p(T_e|U,X,Z,a) p(Z|U,X,a) p(U,X|a) \,.
    \end{align*}
\end{itemize}
Now we show \eqref{eq:cdf_cens_tb}
\begin{align}
    \mathbb{E}\left[ \frac{\mathbb{I}(A=a)q(Z,X,a)\mathbb{I}(T>t)}{P(T_c>t|W,Z,X,a)} \right] & =
    \mathbb{E}\left[ \frac{\mathbb{I}(A=a)q(Z,X,a)}{P(T_c>t|W,Z,X,a)}p(T>t|W,Z,X,a) \right] \notag \\
    & = \mathbb{E}\left[ \frac{\mathbb{I}(A=a)q(Z,X,a)}{P(T_c>t|W,Z,X,a)}p(T_e>t|W,Z,X,a)p(T_c>t|W,Z,X,a) \right] \label{eq:ptc_1} \\
    & = \mathbb{E}[ \mathbb{I}(A=a)q(Z,X,a)p(T_e>t|W,Z,X,a) ] \notag \\
    & = \mathbb{E}[ \mathbb{I}(A=a)q(Z,X,a)\mathbb{I}(T_e>t) ] \notag \\
    & = P(T_e(a)>t) \,. \label{eq:ptc_2}
\end{align}
where in \eqref{eq:ptc_1} we have used $T_e\indep T_c|A,X,W,Z$ and in \eqref{eq:ptc_2} we have used \eqref{eq:do_tb_6}.

\section{Proof of Proposition~\ref{prp:ob}}\label{app:ob}
We show \eqref{eq:do_ob}
\begin{align}
    P(T_e(a)>t) & = \int dX \int dU \ P(T_e(a)>t|U,X)\,p(U,X) \notag \\
    & = \int dX \int dU \ P(T_e>t|U,X,a)\,p(U,X) \label{eq:do_ob_1} \\
    & = \int dX \int dU \int dW \ h(t,W,X,a)\, p(W|U,X)\,p(U,X) \label{eq:do_ob_2} \\
    & = \int dX \int dW \ h(t,W,X,a)\, p(W,X) = \mathbb{E}[h(t,W,X,a)] \,, \label{eq:do_ob_3}
\end{align}
where
\begin{itemize}[leftmargin=4mm,topsep=0mm,itemsep=0mm]
    \item In \eqref{eq:do_ob_1} we have used \eqref{eq:do_eq}.
    \item In \eqref{eq:do_ob_2} we have used
    \begin{align}\label{eq:yuax}
        P(T_e>t|U,X,a) = \int dW \ h(t,W,X,a) p(W|U,X) \,.
    \end{align}
    To show \eqref{eq:yuax}, note that $(W,T_e(a))\indep (A,Z)|U,X$ and $T_e=T_e(A)$ imply $P(T_e>t|U,Z,X,a)=P(T_e>t|U,X,a)$ and $p(W|U,Z,X,a)=p(W|U,X)$, so that
    \begin{align}
        P(T_e>t|Z,X,a) & = \int dU \ P(T_e>t|U,X,a)\, p(U|Z,X,a) \,, \label{eq:pyuxa_1} \\
        \mathbb{E}[h(t,W,X,a)|Z,X,a] & = \int dU \left\{\int dW \ h(t,W,X,a)\, p(W|U,X)\right\} p(U|Z,X,a) \,. \label{eq:pyuxa_2}
    \end{align}
    The left-hand sides of \eqref{eq:pyuxa_1} and \eqref{eq:pyuxa_2} are equal by \eqref{eq:inteq_ob}, hence
    \begin{align}
        \mathbb{E}\left[P(T_e>t|U,X,a)-\int dW \ h(t,W,X,a)\, p(W|U,X)\,\middle|\,Z,X,a\right]=0 \label{eq:pyuxa_3}
    \end{align}
    almost surely, and completeness, i.e., $\mathbb{E}[g_u(U)|Z,X,a]=0$ almost surely only if $g_u(U)=0$ almost surely, yields \eqref{eq:yuax}.
    \item In \eqref{eq:do_ob_3} we have integrated out $U$.
\end{itemize}

Now we show \eqref{eq:cdf_cens_ob}. Since
$\mathbb{I}(T>t)=\mathbb{I}(T_e>t)\mathbb{I}(T_c>t)$, the tower property gives
\begin{align}
    \mathbb{E}\left[
        \frac{\mathbb{I}(T>t)}
        {P(T_c>t|W,Z,X,a)}
        \middle| Z,X,a
    \right]
    &=
    \mathbb{E}\left[
        \mathbb{E}\left[
            \frac{\mathbb{I}(T_e>t)\mathbb{I}(T_c>t)}
            {P(T_c>t|W,Z,X,a)}
            \middle| W,Z,X,a
        \right]
        \middle| Z,X,a
    \right] \notag\\
    &=
    \mathbb{E}\left[
        \frac{
            P(T_e>t|W,Z,X,a)\,
            P(T_c>t|W,Z,X,a)}
        {P(T_c>t|W,Z,X,a)}
        \middle| Z,X,a
    \right] \label{eq:ptc_3}\\
    &=
    \mathbb{E}\left[
        P(T_e>t|W,Z,X,a)
        \middle| Z,X,a
    \right]
    =
    P(T_e>t|Z,X,a) \,.
    \notag
\end{align}
where in \eqref{eq:ptc_3} we have used $T_e\indep T_c|A,X,W,Z$, and censoring positivity ensures that the ratio is well defined.

\section{Proof of Lemma~\ref{lemma:obtb}}\label{app:lemma2}
Let
\[
Y(t)=\frac{\mathbb{I}(T>t)}{P(T_c>t|W,Z,X,A)}.
\]
By \eqref{eq:cdf_cens_ob} and \eqref{eq:inteq_ob},
\[
\mathbb{E}[Y(t)\mid Z,X,a]
=
P(T_e>t\mid Z,X,a)
=
\mathbb{E}[h(t,W,a,X)\mid Z,X,a].
\]
Since $\mathbb{I}(A=a)q(Z,a,X)$ is measurable with respect to $(A,Z,X)$, the tower property gives
\begin{align*}
\mathbb{E}[\mathbb{I}(A=a)q(Z,a,X)Y(t)]
&=
\mathbb{E}\!\left[
\mathbb{I}(A=a)q(Z,a,X)
\mathbb{E}[Y(t)\mid A,Z,X]
\right] \\
&=
\mathbb{E}\!\left[
\mathbb{I}(A=a)q(Z,a,X)
\mathbb{E}[h(t,W,a,X)\mid A,Z,X]
\right] \\
&=
\mathbb{E}[\mathbb{I}(A=a)q(Z,a,X)h(t,W,a,X)] \\
&=
\mathbb{E}[h(t,W,a,X)]
=
P(T_e(a)>t),
\end{align*}
where the penultimate equality is \eqref{eq:align-pop} and the final equality is Proposition~\ref{prp:ob}.

\section{Comparing the two bridge learners}
\label{app:bridge-comparison}
\begin{table}[t]
\centering
% \small
\caption{Comparison of the two outcome bridge learners. Both methods use the same conditioning kernel $k_c$ and Nystr\"om time projection $C_t$; they differ in bridge representation and optimization.}
\label{tab:bridge-comparison}
\footnotesize
\setlength{\tabcolsep}{3pt}
\begin{tabular}{@{}>{\raggedright\arraybackslash}p{0.25\columnwidth}>{\raggedright\arraybackslash}p{0.345\columnwidth}>{\raggedright\arraybackslash}p{0.345\columnwidth}@{}}
\toprule
& \modelnamePMMR{} & \modelnameNMMR{} \\
\midrule
Bridge representation & RKHS (kernel) & Neural network \\
Conditioning kernel $K_c$ & Exact ($n\times n$) & Exact ($B\times B$ per batch) \\
Time structure & Nystr\"om projection $C_t$ & Nystr\"om projection $C_t$ \\
Optimization & Closed-form (linear algebra) & SGD (AdamW) \\
Tuning & ridge $\lambda$; median-heuristic bandwidths & standard neural tuning \\
Suited to & smooth bridges, moderate $n$ & sharper nonlinearity, larger $n$ \\
\bottomrule
\end{tabular}
\end{table}
Table~\ref{tab:bridge-comparison} summarizes the relationship between the two methods.
Both methods minimize the same population-level objective $\mathcal{R}(h)$ (Section~\ref{sec:learning}) and differ in two ways. First, \modelnamePMMR{} represents the bridge in an RKHS and obtains a closed-form solution (Appendix~\ref{app:pmmr-details}), while \modelnameNMMR{} uses a neural network optimized by gradient descent on \eqref{eq:nmmr-projected}. Second, because \modelnamePMMR{}'s RKHS representation has a fixed kernel bandwidth, it works best when the bridge function is relatively smooth; \modelnameNMMR{} can adapt to sharper nonlinearities through its learned features, but requires more data and careful tuning. In practice, these complementary strengths make the choice regime-dependent: \modelnamePMMR{} excels in smoother settings (SEM), while \modelnameNMMR{} is more robust under harder misspecification (Framingham).

\section{\modelnamePMMR{} derivation and implementation details}
\label{app:pmmr-details}

This appendix derives the closed-form Nystr\"om solution for the \modelnamePMMR{} outcome bridge summarized in Section~\ref{sec:learning} (\modelnamePMMR{}: Kernel Bridge Representation).

\subsection{Setup and notation}

Recall that the \modelnamePMMR{} risk is
\[
\hat{\mathcal{R}}_{\mathrm{\modelnamePMMR{}}}(h)
\;=\;
\hat{\mathcal{R}}_r(h)
\;+\;
\lambda\,\|h\|_{\cH}^2,
\]
where $\hat{\mathcal{R}}_r(h)$ is the projected MMR risk~\eqref{eq:common-projected-risk} and the bridge lives in the tensor-product RKHS $\cH=\cH_b\otimes\cH_t$ with bridge kernel $k_b$ on $(W,A,X)$ and time kernel $k_t$ on $\cT$.

We use the following Gram and cross-kernel matrices, all defined on the training data $\{(w_i,a_i,x_i,z_i)\}_{i=1}^n$ and the evaluation grid $\{t_m\}_{m=1}^M$:
\begin{itemize}[leftmargin=1.5em]
\item $K_c\in\mathbb{R}^{n\times n}$: \emph{conditioning} Gram matrix, $[K_c]_{ij}=k_c((a_i,z_i,x_i),\allowbreak(a_j,z_j,x_j))$. Computed exactly.
\item $K_t\in\mathbb{R}^{M\times M}$: \emph{time} Gram matrix, $[K_t]_{m\ell}=k_t(t_m,t_\ell)$.
\item $\Phi_b\in\mathbb{R}^{n\times r_s}$: \emph{bridge} cross-kernel matrix between all $n$ observations and $r_s$ sample landmarks $\{(w_j^{\mathrm{lm}},a_j^{\mathrm{lm}},x_j^{\mathrm{lm}})\}_{j=1}^{r_s}$:
\[
[\Phi_b]_{i,j} \;=\; k_b\!\big((w_i,a_i,x_i),\,(w_j^{\mathrm{lm}},a_j^{\mathrm{lm}},x_j^{\mathrm{lm}})\big).
\]
The sample landmarks are selected by stratified subsampling from the training data (balanced across treatment arms).
\item $G_b\in\mathbb{R}^{r_s\times r_s}$: landmark-to-landmark bridge Gram matrix, $[G_b]_{j,j'}=k_b\!\big((w_j^{\mathrm{lm}},a_j^{\mathrm{lm}},x_j^{\mathrm{lm}}),\allowbreak(w_{j'}^{\mathrm{lm}},a_{j'}^{\mathrm{lm}},x_{j'}^{\mathrm{lm}})\big)$.
\item $C_t\in\mathbb{R}^{K\times r_t}$: \emph{time} cross-kernel matrix between the full grid and $r_t$ uniformly spaced time landmarks $\{t_\ell^{\mathrm{lm}}\}_{\ell=1}^{r_t}$:
\[
[C_t]_{m,\ell} \;=\; k_t(t_m,\,t_\ell^{\mathrm{lm}}).
\]
\item $G_t\in\mathbb{R}^{r_t\times r_t}$: landmark-to-landmark time Gram matrix, $[G_t]_{\ell,\ell'}=k_t(t_\ell^{\mathrm{lm}},t_{\ell'}^{\mathrm{lm}})$.
\item $Y\in\mathbb{R}^{n\times M}$: IPCW pseudo-outcome matrix, $[Y]_{im}=Y_i(t_m)=\mathbb{I}\{\tilde T_i>t_m\}/\hat G(t_m\mid a_i,x_i,w_i,z_i)$.
\end{itemize}

Note that $\Phi_b$ and $G_b$ involve the \emph{bridge} kernel $k_b$ on $(W,A,X)$, while $K_c$ involves the \emph{conditioning} kernel $k_c$ on $(A,Z,X)$. These two kernels operate on different variable sets and serve different roles: $k_b$ defines the function class for the bridge, while $k_c$ defines the test functions that enforce the moment restriction.

\subsection{Representer theorem and Nystr\"om reduction}

By the representer theorem~\cite{scholkopf2001generalized} in $\cH_b\otimes\cH_t$, the minimizer has the form
\[
h(t,w,a,x) \;=\; \sum_{i=1}^n \sum_{m=1}^M \beta_{im}\,k_b\!\big((w,a,x),(w_i,a_i,x_i)\big)\,k_t(t,t_m),
\]
which requires solving for $nM$ coefficients. The Nystr\"om approximation restricts the expansion to the landmark points:
\[
h(t,w,a,x) \;\approx\; \sum_{j=1}^{r_s} \sum_{\ell=1}^{r_t} \alpha_{j\ell}\,k_b\!\big((w,a,x),(w_j^{\mathrm{lm}},a_j^{\mathrm{lm}},x_j^{\mathrm{lm}})\big)\,k_t(t,t_\ell^{\mathrm{lm}}),
\]
reducing the unknowns from $nM$ to $r_s r_t$.

\subsection{Derivation of the Kronecker system}

Substituting the Nystr\"om bridge into the regularized risk and setting the gradient to zero yields the linear system
\begin{equation}
\Big[(\Phi_b^\top K_c\,\Phi_b) \otimes (C_t^\top K_t\,C_t) \;+\; \lambda\,(G_b \otimes G_t)\Big]\,\operatorname{vec}(\alpha) \;=\; \operatorname{vec}\!\big(\Phi_b^\top K_c\,Y\,K_t\,C_t\big),
\label{eq:pmmr-kronecker-app}
\end{equation}
where $\alpha\in\mathbb{R}^{r_s\times r_t}$ and $\operatorname{vec}(\cdot)$ stacks columns. The left-hand side arises from the tensor-product structure of $\cH_b\otimes\cH_t$:
\begin{itemize}[leftmargin=1.5em]
\item The term $\Phi_b^\top K_c\,\Phi_b\in\mathbb{R}^{r_s\times r_s}$ contracts the conditioning kernel against the bridge Nystr\"om features in the sample dimension.
\item The term $C_t^\top K_t\,C_t\in\mathbb{R}^{r_t\times r_t}$ contracts the time kernel against the time Nystr\"om features.
\item The regularizer $\lambda\,(G_b\otimes G_t)$ penalizes the RKHS norm of $h$ expressed through the landmark Gram matrices.
\end{itemize}
The system matrix has size $(r_s r_t)\times(r_s r_t)$. With $r_s=128$ and $r_t=48$, this is $6144\times 6144$, which is solved efficiently by standard dense linear algebra. A small jitter $\epsilon I$ (with $\epsilon=10^{-6}$) is added for numerical stability.

\subsection{Prediction}

Given the solved coefficients $\alpha\in\mathbb{R}^{r_s\times r_t}$, the bridge prediction at a new point $(w,a,x)$ and time $t$ is
\begin{equation}
\hat h(t,w,a,x) \;=\; \mathbf{k}_b(w,a,x)^\top\,\alpha\,\mathbf{k}_t(t),
\label{eq:pmmr-predict}
\end{equation}
where $\mathbf{k}_b(w,a,x)\in\mathbb{R}^{r_s}$ is the vector of bridge-kernel evaluations against the sample landmarks, $[\mathbf{k}_b]_j=k_b((w,a,x),(w_j^{\mathrm{lm}},a_j^{\mathrm{lm}},x_j^{\mathrm{lm}}))$, and $\mathbf{k}_t(t)\in\mathbb{R}^{r_t}$ is the vector of time-kernel evaluations against the time landmarks, $[\mathbf{k}_t]_\ell=k_t(t,t_\ell^{\mathrm{lm}})$. Prediction at all grid points for all observations is therefore $\hat H = \Phi_b\,\alpha\,C_t^\top\in\mathbb{R}^{n\times K}$.

\subsection{Computational complexity}

The dominant costs are:
\begin{itemize}[leftmargin=1.5em]
\item Computing $K_c$: $O(n^2 d_c)$, where $d_c=\dim(Z,X)$.
\item Computing $\Phi_b$: $O(n\,r_s\,d_b)$, where $d_b=\dim(W,X)$.
\item Computing $K_t$ and $C_t$: $O(K^2)$ and $O(K\,r_t)$.
\item Forming and solving the Kronecker system: $O((r_s r_t)^2 n)$ for the matrix products and $O((r_s r_t)^3)$ for the solve.
\end{itemize}
The full $n\times n$ conditioning kernel $K_c$ is the main bottleneck for large $n$, but is computed only once. No iterative optimization or gradient computation is required.

\section{Details about H\'ajek estimator variants}\label{app:hajek}
We report self-normalized H\'ajek variants of the weighted estimators.
We define weights as follows:
\[
w_i^{\mathrm{IPCW}}(a,t)=
\frac{\mathbb{I}\{A_i=a\}\hat q(Z_i,a,X_i)}{\hat G(t\mid A_i,X_i,W_i,Z_i)},
\qquad
w_i^{\mathrm{cov}}(a)=\mathbb{I}\{A_i=a\}\hat q(Z_i,a,X_i),
\]
and $\bar h_a(t)=n^{-1}\sum_{i=1}^n \hat h(t,W_i,a,X_i)$.
Using these weights, the H\'ajek estimators become:
\begin{align}
\hat S_a^{\mathrm{TB-H}}(t)
&=
\frac{\sum_{i=1}^n w_i^{\mathrm{IPCW}}(a,t)\mathbb{I}\{\tilde T_i>t\}}
{\sum_{i=1}^n w_i^{\mathrm{cov}}(a)},
\label{eq:pipw-h}\\
\hat S_a^{\mathrm{DR-H}}(t)
&=
\bar h_a(t)+
\frac{\sum_{i=1}^n w_i^{\mathrm{IPCW}}(a,t)\mathbb{I}\{\tilde T_i>t\}}
{\sum_{i=1}^n w_i^{\mathrm{cov}}(a)}
-
\frac{\sum_{i=1}^n w_i^{\mathrm{cov}}(a)\hat h(t,W_i,a,X_i)}
{\sum_{i=1}^n w_i^{\mathrm{cov}}(a)}.
\label{eq:pdr-h}
\end{align}
 Both terms are normalized by the treatment weights, whose mean estimates $\E[\mathbb{I}\{A=a\}q(Z,a,X)]=1$ (take $f\equiv1$ in \eqref{eq:tb-moment}). The censoring weight $1/\hat G$ is left out of the normalizing sum because it is offset by the censoring indicator contained in $\mathbb{I}\{\tilde T_i>t\}$, which appears only in the numerator; normalizing by $\sum_i w_i^{\mathrm{IPCW}}(a,t)$ would target $S_a(t)/\E[\mathbb{I}\{A=a\}q(Z,a,X)/G(t\mid A,X,W,Z)]$ instead of $S_a(t)$ whenever there is censoring before $t$.  With $\hat m_a=n^{-1}\sum_{i=1}^n w_i^{\mathrm{cov}}(a)$, \eqref{eq:pipw-h} and \eqref{eq:pdr-h} equal $\hat S_a^{\mathrm{TB}}(t)/\hat m_a$ and $\hat S_a^{\mathrm{OB}}(t)+\{\hat S_a^{\mathrm{DR}}(t)-\hat S_a^{\mathrm{OB}}(t)\}/\hat m_a$, so the theory for the standard estimators extends directly. If $\hat q$ converges to a solution of \eqref{eq:inteq_tb}, then $\hat m_a\to1$; if $\hat h$ converges to a solution of \eqref{eq:inteq_ob}, then $\hat S_a^{\mathrm{DR}}(t)-\hat S_a^{\mathrm{OB}}(t)\to0$. Provided $\hat m_a$ stays bounded away from zero, the H\'ajek DR estimator is therefore consistent under the conditions of Theorem~\ref{thm:dr}, the H\'ajek TB estimator is consistent whenever the standard one is, and asymptotic normality carries over by the delta method wherever it holds jointly for the standard estimators and $\hat m_a$.

\section{Compared approaches and metrics}
\label{app:compared-approaches-metrics}
We conduct experiments and compare against the parametric proximal bridge family of \citet{ying2022proximal}. Their survival bridge restricts $h(t,w,a,x)$ to a low-dimensional exponential form that is linear in $(a,w,x)$ and quadratic in time, together with a matching low-dimensional treatment bridge, and fits the bridge parameters by solving the corresponding sample moment equations. For PIPW and PDR, we use \citet{ying2022proximal}'s same formulation as in their paper. Since \citet{ying2022proximal} does not define POR, we use the following, which could be seen as an extension of \citet{cui2024semiparametric} for time-to-event outcomes. In particular,
$
\hat S_a^{\mathrm{POR}}(t)=\frac{1}{n}\sum_{i=1}^n \hat h_t^{\mathrm{Ying}}(W_i,a,X_i),
$
with $\hat\psi^{\mathrm{POR}}(t)=\hat S_1^{\mathrm{POR}}(t)-\hat S_0^{\mathrm{POR}}(t)$. The advantage of this parametric model is its simplicity and ease of implementation. It is especially informative on SEM, where the simulation is generated from a data-generating mechanism that is favorable to its functional form.

We also compare against three standard weighted Cox proportional hazards baselines. Let $\hat e_i=\hat \Prob(A_i=1\mid V_i)$ be a logistic propensity estimate fit from the covariates $V_i$. We then fit a Cox model with treatment $A$ as the sole regression covariate under three weighting rules~\citep{buchanan2014worth, rosenbaum1983central, schemper2009estimation}:
$
\omega_i^{\mathrm{Unif.}}=1,
\omega_i^{\mathrm{IPW}}=\frac{A_i}{\hat e_i}+\frac{1-A_i}{1-\hat e_i},
\omega_i^{\mathrm{OW}}=A_i(1-\hat e_i)+(1-A_i)\hat e_i.
$
Uniform is the unweighted Cox baseline, IPW is the usual inverse-probability weighting estimator, and OW emphasizes the overlap population \citep{rosenbaum1983central,cao2009improving,li2018balancing}. Appendix \ref{app:baselines} gives the weighted Cox partial likelihood. These baselines are useful precisely because they do \emph{not} solve proximal bridge equations. Across experiments we report average absolute bias, MAE, RMSE, and,
empirical coverage; formal definitions appear in Appendix \ref{app:metrics}. Avg.\ abs.\ bias captures systematic error across horizons, MAE captures typical absolute error, RMSE upweights occasional large failures, which is important for unstable weighted estimators, and coverage is defined in Appendix~\ref{app:metrics}; it measures the calibration of 95\% intervals only for the SEM proximal bridge estimators.

\section{Dataset details used in the Experiments}
\label{app:data}
\subsection{SEM simulation design}
\label{app:simulation-details}

For the SEM benchmark, we follow the simulation setup of \citet{ying2022proximal}. Under the notation of Section~\ref{sec:background}, we generate i.i.d.\ samples
\(
O_i=(T_i,A_i,X_i,Z_i,W_i)
\)
through the following data-generating mechanism. First, we generate the latent
covariate \(X\), unmeasured confounder \(U\), treatment \(A\), treatment-confounding
proxy \(Z\), outcome-confounding proxy \(W\), latent event time \(T^\star\), and
censoring time \(C\) as
\begin{align}
X &\sim \max\!\bigl(\operatorname{Normal}(1.1,0.75^2),\,0\bigr), \\
U &\sim \max\!\bigl(\operatorname{Normal}(1.1,0.75^2),\,0\bigr), \\
\mathbb{P}(A=1 \mid X,U)
&= \frac{1}{1+\exp\!\bigl(-(0.3+0.4X-0.6U)\bigr)}, \\
Z &\sim \operatorname{Normal}(-0.2-0.3X+0.65U,\,0.5^2), \\
W &\sim \operatorname{Normal}(-0.6+0.4X+0.65U,\,0.5^2), \\
T^\star &\sim \operatorname{Exponential}(0.1+0.6A+0.25X+0.5U), \\
C &\sim \min\!\bigl(\operatorname{Exponential}(0.2),\,2\bigr).
\end{align}
We then observe
\begin{align}
T &= \min(T_e,T_c).%
\end{align}

The censoring variable \(T_c\) is generated independently of
\((A,X,Z,W,U,T_e)\), yielding approximately 22\% censoring.

\subsection{Framingham cohort study}
\label{app:fram}

The Framingham benchmark uses the Offspring cohort of the Framingham Heart Study \citep{benjamin1994independent}. We study statin exposure and cardiovascular-event risk. Variables are assigned to blocks by the proxy-allocation strategy of \citet{meng2025coupling}, inspired by \citet{tchetgen2020introduction} and also used by \citet{cui2024semiparametric}: each of the 32 candidate covariates from Offspring Exam~6 receives a treatment-relevance score (absolute logistic-regression coefficient for $A$) and an outcome-relevance score (absolute Cox coefficient) and is allocated to the block for which it is more relevant. All 32 candidates are allocated to the two proxy blocks, so the $X$ block is empty. The treatment-side proxies $Z$ are \texttt{mi\_hx6}, \texttt{dbp62}, \texttt{bpmeds6}, \texttt{hdl52}, \texttt{dbp61}, \texttt{bmi62}, \texttt{hdl51}, \texttt{bmi61}, \texttt{bmi6}, \texttt{chol52}, \texttt{dbp6}, \texttt{sbp61}, \texttt{chol51}, \texttt{chol5}, \texttt{sbp6}, and \texttt{trigly5}; the outcome-side proxies $W$ are \texttt{smoke6}, \texttt{female}, \texttt{ascvd\_hx6}, \texttt{diab6}, \texttt{pad\_hx6}, \texttt{gluc52}, \texttt{stk\_hx6}, \texttt{age62}, \texttt{sbp62}, \texttt{gluc51}, \texttt{age61}, \texttt{age6}, \texttt{hdl5}, \texttt{gluc5}, \texttt{trigly52}, and \texttt{trigly51}. Plausible unmeasured confounders in this statin setting are overall metabolic health, health-seeking behavior, and frailty. Treatment-side proxies, such as prior myocardial infarction, blood-pressure medication, and lipid measurements, drive the prescribing decision without directly affecting the outcome given the latent confounder, whereas outcome-side proxies are physiological measurements affected by the confounder but not directly by treatment. Each of the 30 runs uses the full pooled cohort without resampling participants (no bootstrap); runs differ only in the random seed, which changes the internal train/validation split and the stochastic fitting (initialization and minibatch optimization). The benchmark evaluates the resulting survival-curve estimates at normalized horizons $\{0.3,0.5,0.7\}$, corresponding to $\{2184,3640,5095\}$ days on the pooled follow-up scale.
The external benchmark is the constant hazard ratio $0.75$ reported by the external randomized trial~\citep{yusuf2016cholesterol}, which enrolled a different population; we therefore use it as an external reference rather than as the true effect in the cohort.

\subsection{ACTG data generation process}
\label{app:actg-dgp}

The proximal semi-synthetic ACTG dataset keeps the original ACTG175 baseline covariates from \citet{hammer1996trial} fixed and regenerates only the latent, proxy, treatment, event-time, and censoring variables. We define
\[
\Std(v)=\frac{v-\bar v}{\mathrm{sd}(v)}.
\]
Using standardized age, hemoglobin, symptom burden, weight, Karnofsky score, and log-CD4/log-CD8 features, the generator forms three nonlinear observed summaries:
\begin{align}
x_i^{(\mathrm{trt})}
&=
\Std\Big(
0.55\,\widetilde{\mathrm{age}}_i\widetilde{\mathrm{symptom}}_i
-0.45\,\widetilde{\log(1+\mathrm{cd40})}_i\widetilde{\mathrm{hemo}}_i
+0.35\sin(1.3\,\widetilde{\mathrm{age}}_i) \notag\\
&\hspace{0.cm}
+0.30\{\widetilde{\log(1+\mathrm{cd80})}_i^2-\E[\widetilde{\log(1+\mathrm{cd80})}^2]\}
-0.25\,\widetilde{\mathrm{wtkg}}_i\widetilde{\mathrm{karnof}}_i
\Big), \notag\\
x_i^{(\mathrm{out})}
&=
\Std\Big(
0.60\{\widetilde{\mathrm{age}}_i^2-\E[\widetilde{\mathrm{age}}^2]\}
+0.45\,\widetilde{\log(1+\mathrm{cd80})}_i\widetilde{\mathrm{symptom}}_i
-0.40\tanh(\widetilde{\mathrm{hemo}}_i) \notag\\
&\hspace{0.cm}
+0.30\{\widetilde{\mathrm{wtkg}}_i^2-\E[\widetilde{\mathrm{wtkg}}^2]\}
-0.25\,\widetilde{\log(1+\mathrm{cd40})}_i\widetilde{\mathrm{preanti}}_i
+0.25\sin(\widetilde{\mathrm{karnof}}_i)
\Big), \notag\\
x_i^{(\mathrm{shared})}
&=
\Std\!\Big(
0.50\,x_i^{(\mathrm{trt})}+0.50\,x_i^{(\mathrm{out})}
+0.20\,\widetilde{\mathrm{age}}_i\widetilde{\log(1+\mathrm{cd80})}_i
\Big).
\label{eq:actg-observed-summaries}
\end{align}

The latent confounder is generated from independent Gaussian seeds
$U_i^{\mathrm{raw}},U_i^{\mathrm{aux}}\stackrel{\mathrm{iid}}{\sim}N(0,1)$ via
\begin{align}
U_i
&=
\Std\!\Big(
U_i^{\mathrm{raw}}
+0.40\,\mathrm{sign}(U_i^{\mathrm{aux}})
\sqrt{|U_i^{\mathrm{aux}}|+10^{-6}}
\Big), \notag\\
U_i^{\mathrm{phase}}
&=
\Std\!\left(0.70\,U_i+0.30\sin(1.2\,U_i^{\mathrm{aux}})\right),
\qquad
U_i^{(2)}=U_i^2-1,
\label{eq:actg-latent}
\end{align}
and a shared severity score
\begin{align}
\mathrm{sev}_i
&=
1.20\,U_i+0.20\,U_i^{(2)}
-0.50\,\widetilde{\log(1+\mathrm{cd40})}_i
-0.30\,\widetilde{\log(1+\mathrm{cd80})}_i
+0.20\,\widetilde{\mathrm{age}}_i \notag\\
&\quad
+0.25\,\widetilde{\mathrm{symptom}}_i
-0.15\,\widetilde{\mathrm{hemo}}_i
+0.45\,x_i^{(\mathrm{shared})}\notag\\
&\quad
+0.30\sin(1.5\,x_i^{(\mathrm{shared})})
+0.20\,\widetilde{\mathrm{age}}_i\widetilde{\log(1+\mathrm{cd80})}_i,
\notag\\
S_i
&=
\Std\!\Big(
0.35\,\mathrm{sev}_i
+0.40\tanh(1.1\,\mathrm{sev}_i)
+0.25\sin(1.7\,\mathrm{sev}_i) \notag\\
&\quad
+0.20\max(\mathrm{sev}_i-0.15,0)
\Big).
\label{eq:actg-severity}
\end{align}

The role-separated proxies are
\begin{align}
z_{1i}
&=
2.0\,\sin(1.2\,S_i+0.35\,U_i^{\mathrm{phase}})
+0.55\,x_i^{(\mathrm{trt})}
+0.30\tanh(U_i)+\varepsilon_i^{(z1)}, \notag\\
z_{2i}
&=
1.8\,\cos(1.35\,S_i-0.25\,x_i^{(\mathrm{shared})})
+0.30\,\widetilde{\mathrm{age}}_i
+0.35\,x_i^{(\mathrm{shared})}
-0.20\tanh(U_i^{\mathrm{phase}})
+\varepsilon_i^{(z2)}, \notag\\
Z_i
&=
\Std\!\Big(
0.90\tanh(z_{1i})-0.60z_{2i}
+0.35\sin(0.8z_{1i}z_{2i})
+0.25\{z_{2i}^2-\E[z_2^2]\}
\Big),
\label{eq:actg-z}
\end{align}
with $\varepsilon_i^{(z1)},\varepsilon_i^{(z2)}\sim N(0,0.35^2)$, and
\begin{align}
w_{1i}
&=
1.40\,S_i-0.25\,\widetilde{\mathrm{hemo}}_i
+0.15\,\widetilde{\mathrm{symptom}}_i
+0.60\,x_i^{(\mathrm{out})}
+0.35\sin(1.1U_i^{\mathrm{phase}})
+\varepsilon_i^{(w1)}, \notag\\
w_{2i}
&=
(S_i^2-\E[S^2])
+0.20\,\widetilde{\log(1+\mathrm{cd80})}_i
+0.40\,x_i^{(\mathrm{out})}\tanh(S_i)
+0.25\,x_i^{(\mathrm{shared})}
+\varepsilon_i^{(w2)}, \notag\\
W_i
&=
\Std\!\Big(
0.75\tanh(w_{1i})+0.35w_{2i}
+0.30\sin(w_{1i}-0.8w_{2i})
+0.20\{w_{1i}^2-\E[w_1^2]\}
\Big),
\label{eq:actg-w}
\end{align}
with $\varepsilon_i^{(w1)}\sim N(0,0.45^2)$ and $\varepsilon_i^{(w2)}\sim N(0,0.55^2)$.

The transformed proxy features used later are
\begin{align}
Z_i^{\tanh}&=\tanh(Z_i), &
Z_i^{\mathrm{wave}}&=\sin(1.8Z_i)+0.25\cos(1.1z_{2i}), \notag\\
Z_i^{\mathrm{gate}}
&=
\begin{cases}
Z_i^2-\E[Z^2], & Z_i\ge 0,\\
-0.60|Z_i|, & Z_i<0,
\end{cases}
&
W_i^{\tanh}&=\tanh(W_i), \notag\\
W_i^{\mathrm{wave}}&=\sin(1.4W_i)+0.35\cos(0.9w_{2i}),\qquad
&
W_i^{\mathrm{gate}}
&=
\begin{cases}
(W_i-0.15)^2, & W_i\ge 0.15,\\
-0.45\,(W_i+0.15)^2, & W_i<0.15.
\end{cases}
\label{eq:actg-proxy-transforms}
\end{align}

Treatment is drawn from
\begin{align}
\eta_i^{(A)}
&=
-0.15\,\widetilde{\mathrm{age}}_i
-0.20\,\widetilde{\log(1+\mathrm{cd40})}_i
-0.85\,U_i
-0.15\,U_i^{(2)}
+0.45\,x_i^{(\mathrm{trt})} \notag\\
&\quad
+0.18\,x_i^{(\mathrm{trt})}Z_i^{\tanh}
+0.90\,Z_i^{\mathrm{wave}}\notag\\
&\quad
+0.20\,Z_i^{\mathrm{gate}}
+0.20\sin(U_i^{\mathrm{phase}}Z_i)
-0.18\,\mathbb{I}\{Z_i<-0.35\},
\notag\\
\pi_i
&=
0.12+0.76\,\sigmoid(0.35+\eta_i^{(A)}),
\notag\\
A_i&\sim \mathrm{Bernoulli}(\pi_i).
\label{eq:actg-treatment}
\end{align}

Potential event times are generated on a 240-point grid. Let
$s_j=(j-1)/239$ and
\[
\tau_j=\max\{1,\;1800(0.02+0.98s_j^{1.15})\},
\qquad j=1,\dots,240.
\]
With $L_i=\Std(X_i^\top(0.60\beta_0+0.40\beta_1))$,
\begin{align}
r(s)&=-6.10+4.20s-1.40(s-0.58)^2+0.35\sin(4\pi s), \notag\\
\ell(s)&=0.55+0.90s+0.25\sin(2\pi s), \qquad
\ell_{\mathrm{late}}(s)=\max(s-0.45,0), \notag\\
R_i
&=
0.25\tanh(1.15U_i)+0.12U_i^{(2)}+0.11W_i^{\mathrm{wave}}
+0.18W_i^{\mathrm{gate}}+0.15\tanh(U_iW_i^{\tanh}) \notag\\
&\quad
+0.35x_i^{(\mathrm{out})}
+0.20x_i^{(\mathrm{out})}W_i^{\tanh}
+0.30L_i, \notag\\
H_i
&=
0.60\tanh(0.9W_i+0.50x_i^{(\mathrm{out})}-0.35U_i)
+0.25\sin(1.1W_i^{\mathrm{wave}})
-0.20\tanh(U_i^{\mathrm{phase}}), \notag\\
g(s)
&=
0.10+0.95\,\sigmoid(12(s-0.55))-0.15\,\sigmoid(12(s-0.20)), \notag\\
\Omega_i(s)
&=
0.18\tanh(W_i^{\mathrm{gate}})\sin(3\pi s).
\label{eq:actg-hazard-components}
\end{align}
The arm-specific hazard logits and discrete hazards are
\begin{align}
\eta_{0i}(s)
&=
r(s)+R_i\ell(s)+\max(W_i-0.15,0)\ell_{\mathrm{late}}(s)+\Omega_i(s), \notag\\
\eta_{1i}(s)
&=
\eta_{0i}(s)-H_i g(s)+0.12\,x_i^{(\mathrm{out})}\cos(2.5\pi s), \notag\\
h_{ai}(s)
&=
\mathrm{clip}\bigl(\sigmoid(\eta_{ai}(s)),10^{-4},0.35\bigr),
\qquad a\in\{0,1\}.
\label{eq:actg-discrete-hazard}
\end{align}
With independent uniforms $V_{ai}\sim \mathrm{Unif}(0,1)$ and
$F_{ai}(\tau_j)=1-\prod_{k\le j}(1-h_{ai}(s_k))$, the raw event times are
\begin{equation}
T_{ai}^{\mathrm{raw}}
=
\begin{cases}
\min\{\tau_j:F_{ai}(\tau_j)\ge V_{ai}\}, & \text{if }F_{ai}(\tau_{240})\ge V_{ai},\\
\tau_{240}+\dfrac{180}{\max\{h_{ai}(s_{240}),10^{-3}\}}, & \text{otherwise}.
\end{cases}
\label{eq:actg-raw-event}
\end{equation}

The proxy-dependent warp is
\begin{align}
\omega_i
&=
\Std\Big(
0.95\tanh(w_{1i})\max(w_{2i}-0.05,0)
+0.65\sin(1.2w_{1i}-0.8w_{2i}) \notag\\
&\hspace{1.6cm}
+0.40x_i^{(\mathrm{out})}\tanh(S_i)-0.35U_i^{\mathrm{phase}}
\Big), \notag\\
G_i
&=
\mathbb{I}\{w_{1i}w_{2i}+0.60\sin(w_{1i})+0.20x_i^{(\mathrm{out})}>0.12\}, \notag\\
K_i&=390+200\,\sigmoid(\omega_i)+220\,G_i.
\label{eq:actg-warp-score}
\end{align}
The final potential event times are
\begin{align}
T_{0i}
&=
\begin{cases}
0.72\,T_{0i}^{\mathrm{raw}}, & T_{0i}^{\mathrm{raw}}\le K_i,\; G_i=1,\\
0.64\,T_{0i}^{\mathrm{raw}}, & T_{0i}^{\mathrm{raw}}\le K_i,\; G_i=0,\\
0.72\,K_i+0.38(T_{0i}^{\mathrm{raw}}-K_i), & T_{0i}^{\mathrm{raw}}>K_i,\; G_i=1,\\
0.64\,K_i+0.22(T_{0i}^{\mathrm{raw}}-K_i), & T_{0i}^{\mathrm{raw}}>K_i,\; G_i=0,
\end{cases}
\notag\\
T_{1i}
&=
\begin{cases}
1.28\,T_{1i}^{\mathrm{raw}}+70\,\sigmoid(\omega_i), & T_{1i}^{\mathrm{raw}}\le K_i,\; G_i=1,\\
0.86\,T_{1i}^{\mathrm{raw}}+5\,\sigmoid(\omega_i), & T_{1i}^{\mathrm{raw}}\le K_i,\; G_i=0,\\
K_i+2.15(T_{1i}^{\mathrm{raw}}-K_i)+340\,\sigmoid(\omega_i), & T_{1i}^{\mathrm{raw}}>K_i,\; G_i=1,\\
K_i+1.48(T_{1i}^{\mathrm{raw}}-K_i)+150\,\sigmoid(\omega_i), & T_{1i}^{\mathrm{raw}}>K_i,\; G_i=0.
\end{cases}
\label{eq:actg-warped-time}
\end{align}

Finally, censoring is generated by
\begin{align}
\mu_i^{(C)}
&=
\log(1000)+0.10x_i^{(\mathrm{shared})}-0.12\tanh(Z_i)+0.08|W_i^{\mathrm{wave}}|,
\notag\\
C_i
&=
\max\Big\{\exp\big(\mu_i^{(C)}+0.28\,\xi_i\big),\,1\Big\},
\qquad \xi_i\sim N(0,1),
\label{eq:actg-censor}
\end{align}
and the observed factual/counterfactual data are
\begin{align}
T_i^{\mathrm{f}}&=A_iT_{1i}+(1-A_i)T_{0i}, &
T_i^{\mathrm{cf}}&=A_iT_{0i}+(1-A_i)T_{1i}, \notag\\
Y_i^{\mathrm{f}}&=\min(T_i^{\mathrm{f}},C_i), &
E_i^{\mathrm{f}}&=\mathbb{I}\{T_i^{\mathrm{f}}\le C_i\},
\label{eq:actg-factual}
\end{align}
with the analogous counterfactual pair $(Y_i^{\mathrm{cf}},E_i^{\mathrm{cf}})$.

\section{Evaluation metrics}
\label{app:metrics}

Let $\Theta$ denote the set of reported horizons and let $\hat\theta_{m,r}(t)$ be the estimate produced by method $m$ on replicate $r$ at horizon $t$, with benchmark target $\theta_r^\star(t)$. For survival-curve experiments, $\theta_r^\star(t)$ is the target treatment effect $\psi_r(t)=S_{1,r}(t)-S_{0,r}(t)$; for Framingham it is the external benchmark hazard-ratio described below. With $R$ replicates,
\begin{align}
\mathrm{Bias}_m(t) &= \frac{1}{R}\sum_{r=1}^R \big(\hat\theta_{m,r}(t)-\theta_r^\star(t)\big),\\
\mathrm{MAE}_m(t) &= \frac{1}{R}\sum_{r=1}^R \big|\hat\theta_{m,r}(t)-\theta_r^\star(t)\big|,\\
\mathrm{RMSE}_m(t) &= \left\{\frac{1}{R}\sum_{r=1}^R \big(\hat\theta_{m,r}(t)-\theta_r^\star(t)\big)^2\right\}^{1/2},\\
\mathrm{Cov}_m(t) &= \frac{1}{R}\sum_{r=1}^R \mathbb{I}\!\left\{\theta_r^\star(t)\in \mathrm{CI}_{m,r}(t)\right\}.
\end{align}
The interval $\mathrm{CI}_{m,r}(t)$ is dataset-specific. For the SEM proximal bridge estimators (POR, PIPW, PDR, and OB, TB, DR), $\mathrm{CI}_{m,r}(t)=\hat\theta_{m,r}(t)\pm1.96\,\widehat{\mathrm{SE}}_{m,r}(t)$ is a pointwise Wald interval whose standard error is computed from the estimated influence function in replicate $r$, so $\mathrm{Cov}_m(t)$ is the usual empirical coverage over $R=500$ replicates. The SEM Cox baselines do not retain separate interval outputs, so their coverage values are reconstructed from the replicate mean and standard deviation as a normal-approximation summary. For ACTG and Framingham, per-fit standard errors are not used: $\mathrm{CI}_{m,r}(t)=\hat\theta_{m,r}(t)\pm1.96\,\hat\sigma_m(t)$, where $\hat\sigma_m(t)$ is the standard deviation of $\hat\theta_{m,1}(t),\dots,\hat\theta_{m,R}(t)$. This \emph{replicate-dispersion coverage} indicates whether the bias is small relative to the Monte Carlo spread; it is not the coverage of a confidence interval that could be computed from a single dataset. For Framingham, the $R=30$ runs share one cohort and differ only in the train/validation split and the stochastic fitting, so $\hat\sigma_m(t)$ reflects computational variability only, and $\mathrm{Cov}_m(t)$ is the rate at which the external benchmark lies within these bands. With $R$ replicates, the Monte Carlo standard error of a coverage near 0.95 is about $\{0.95\cdot0.05/R\}^{1/2}$: 0.010 for $R=500$ and 0.040 for $R=30$.
The tables in the main text report averages over the reported horizons, for example
\[
\mathrm{AvgAbsBias}_m = \frac{1}{|\Theta|}\sum_{t\in\Theta} |\mathrm{Bias}_m(t)|.
\]
MAE and RMSE play different roles in the empirical summaries: MAE reports the typical absolute error at a horizon, whereas RMSE is more sensitive to occasional large failures and therefore highlights instability in weighted estimators.

For the Framingham benchmark we first convert each estimated pair of survival curves $(\hat S_{1,m,r},\hat S_{0,m,r})$ into the hazard-ratios.%
Let $\hat m_{a,m,r}$ be the slope from regressing $\hat S_{a,m,r}(t)$ on the evaluation grid. Then
\begin{equation}
\widehat{\mathrm{HR}}_{m,r}(t)
=
\frac{\hat S_{0,m,r}(t)}{\hat S_{1,m,r}(t)+\varepsilon}
\cdot
\frac{\hat m_{1,m,r}}{\hat m_{0,m,r}+\varepsilon},
\qquad \varepsilon=10^{-8},
\end{equation}
and the same bias, MAE, RMSE, and coverage formulas are applied to this scalar target.

\section{Weighted Cox baselines}
\label{app:baselines}

The Cox baselines estimate a propensity score $\hat e_i=\hat \Prob(A_i=1\mid X_i)$ by logistic regression and then maximize a weighted partial likelihood with treatment as the sole Cox covariate:
\begin{equation}
\mathcal{L}(\gamma)
=
\prod_{m:\delta_m=1}
\left(
\frac{\exp(x_m\gamma)}
{\sum_{n:\tilde T_n\ge \tilde T_m}\omega_n \exp(x_n\gamma)}
\right)^{\omega_m}.
\label{eq:weighted-cox}
\end{equation}
Here $\delta_m=1$ indicates an observed event and $\tilde T_m=\min(T_m^\star,C_m)$ is the observed follow-up time. The three weighting rules are
\begin{align}
\omega_i^{\mathrm{Uniform}}&=1,\\
\omega_i^{\mathrm{IPW}}&=\frac{A_i}{\hat e_i}+\frac{1-A_i}{1-\hat e_i},\\
\omega_i^{\mathrm{OW}}&=A_i(1-\hat e_i)+(1-A_i)\hat e_i.
\end{align}
These baselines are standard balancing-weight survival estimators, but unlike the proximal methods they do not attempt to solve the bridge equations \eqref{eq:inteq_ob} or \eqref{eq:tb-moment}.

\section{Detailed Results}
\label{app:detailed-results}

The main-text tables and figure report averages over the displayed horizons. Tables \ref{tab:actg-detailed}, \ref{tab:framingham-detailed}, and \ref{tab:sem-detailed} expand those summaries into per-horizon breakdowns. All three datasets report bias, MAE, RMSE, and coverage.
Bias is signed in every per-horizon column, and the final average columns are averages over the displayed horizons; in Table~\ref{tab:sem-detailed}, the average Bias column is the average absolute bias.

\begin{table}[t]
\caption{Detailed ACTG results for Table \ref{tab:sem-actg}. The displayed ACTG horizons are $\{201,401,602,802\}$ days. Bias is signed, and the final average columns are simple averages over these horizons. Cov is replicate-dispersion coverage over 30 datasets (Appendix~\ref{app:metrics}).}
\label{tab:actg-detailed}
\centering
% \scriptsize
\resizebox{\textwidth}{!}{%
\begin{tabular}{lcccccccccccccccccccc}
\toprule
& \multicolumn{4}{c}{201 d} & \multicolumn{4}{c}{401 d} & \multicolumn{4}{c}{602 d} & \multicolumn{4}{c}{802 d} & \multicolumn{4}{c}{Avg.} \\
Method & Bias & MAE & RMSE & Cov & Bias & MAE & RMSE & Cov & Bias & MAE & RMSE & Cov & Bias & MAE & RMSE & Cov & Bias & MAE & RMSE & Cov \\
\midrule
\multicolumn{21}{l}{\emph{Weighted Cox baselines}} \\
Uniform & 0.081 & 0.081 & 0.084 & 0.133 & 0.120 & 0.120 & 0.126 & 0.267 & -0.028 & 0.029 & 0.040 & 0.667 & -0.058 & 0.058 & 0.060 & 0.300 & 0.029 & 0.072 & 0.078 & 0.342 \\
IPW & 0.084 & 0.084 & 0.086 & 0.067 & 0.128 & 0.128 & 0.134 & 0.267 & -0.014 & 0.026 & 0.033 & 0.767 & -0.047 & 0.047 & 0.049 & 0.600 & 0.038 & 0.071 & 0.076 & 0.425 \\
OW & 0.083 & 0.083 & 0.085 & 0.067 & 0.127 & 0.127 & 0.132 & 0.267 & -0.014 & 0.026 & 0.033 & 0.767 & -0.047 & 0.047 & 0.049 & 0.600 & 0.037 & 0.071 & 0.075 & 0.425 \\
\addlinespace
\multicolumn{21}{l}{\emph{Parametric proximal bridge \citep{ying2022proximal}}} \\
POR & 0.115 & 0.118 & 0.129 & 0.433 & 0.001 & 0.107 & 0.127 & 0.833 & -0.239 & 0.239 & 0.258 & 0.233 & -0.270 & 0.270 & 0.278 & 0.000 & -0.098 & 0.184 & 0.198 & 0.375 \\
PIPW & 0.018 & 0.030 & 0.037 & 0.933 & -0.008 & 0.026 & 0.032 & 1.000 & -0.036 & 0.036 & 0.040 & 1.000 & -0.038 & 0.039 & 0.043 & 1.000 & -0.016 & 0.033 & 0.038 & 0.983 \\
PDR & 0.015 & 0.017 & 0.021 & 0.900 & 0.003 & 0.024 & 0.026 & 1.000 & -0.019 & 0.023 & 0.026 & 1.000 & -0.020 & 0.023 & 0.027 & 1.000 & -0.005 & 0.022 & 0.025 & 0.975 \\
\addlinespace
\multicolumn{21}{l}{\emph{Flexible proximal bridges}} \\
OB-\modelnameNMMR{} & 0.035 & 0.041 & 0.054 & 0.900 & -0.017 & 0.055 & 0.069 & 0.900 & -0.005 & 0.028 & 0.035 & 1.000 & 0.037 & 0.049 & 0.057 & 1.000 & 0.013 & 0.043 & 0.054 & 0.950 \\
TB-\modelnameNMMR{} & 0.031 & 0.043 & 0.056 & 0.900 & 0.016 & 0.032 & 0.041 & 1.000 & 0.006 & 0.019 & 0.024 & 1.000 & 0.000 & 0.016 & 0.021 & 1.000 & 0.013 & 0.027 & 0.036 & 0.975 \\
DR-\modelnameNMMR{} & 0.009 & 0.015 & 0.017 & 0.967 & -0.002 & 0.023 & 0.027 & 1.000 & -0.007 & 0.014 & 0.018 & 1.000 & -0.007 & 0.013 & 0.017 & 1.000 & -0.002 & 0.016 & 0.020 & 0.992 \\
OB-\modelnamePMMR{} & -0.028 & 0.028 & 0.032 & 0.767 & 0.004 & 0.030 & 0.037 & 1.000 & 0.004 & 0.012 & 0.015 & 1.000 & 0.002 & 0.011 & 0.014 & 1.000 & -0.005 & 0.020 & 0.024 & 0.942 \\
TB-\modelnamePMMR{} & 0.039 & 0.048 & 0.062 & 0.833 & 0.022 & 0.035 & 0.044 & 1.000 & 0.010 & 0.020 & 0.025 & 0.967 & 0.003 & 0.015 & 0.020 & 1.000 & 0.019 & 0.030 & 0.038 & 0.950 \\
DR-\modelnamePMMR{} & 0.006 & 0.013 & 0.016 & 1.000 & 0.002 & 0.019 & 0.024 & 1.000 & 0.001 & 0.011 & 0.013 & 1.000 & -0.000 & 0.011 & 0.014 & 1.000 & 0.002 & 0.014 & 0.017 & 1.000 \\
\bottomrule
\end{tabular}%
}
\end{table}

\begin{table}[t]
\caption{Detailed Framingham results for the standard (non-H\'ajek) estimators. The displayed horizons are $\{2184,3640,5095\}$ days. Bias is signed, and the final average columns are simple averages over these horizons;  Table~\ref{tab:hajek-ablation-full} reports the H\'ajek variants of the weighted estimators. Cov is the rate at which the external benchmark lies within 1.96 across-run standard deviations of each run's estimate, over 30 runs on a fixed cohort (Appendix~\ref{app:metrics}); it is not the coverage of a confidence interval. Runs whose hazard ratio is not finite (an estimated survival curve projected to one) are excluded: 7, 7, and 3 of the 30 runs for PIPW and 1, 1, and 0 for TB-\modelnamePMMR{}.}
\label{tab:framingham-detailed}
\centering
% \scriptsize
\resizebox{\textwidth}{!}{%
\begin{tabular}{lcccccccccccccccc}
\toprule
& \multicolumn{4}{c}{2184 d} & \multicolumn{4}{c}{3640 d} & \multicolumn{4}{c}{5095 d} & \multicolumn{4}{c}{Avg.} \\
Method & Bias & MAE & RMSE & Cov & Bias & MAE & RMSE & Cov & Bias & MAE & RMSE & Cov & Bias & MAE & RMSE & Cov \\
\midrule
\multicolumn{17}{l}{\emph{Weighted Cox baselines}} \\
Uniform & 1.026 & 1.026 & 1.045 & 0.000 & 1.026 & 1.026 & 1.045 & 0.000 & 1.026 & 1.026 & 1.045 & 0.000 & 1.026 & 1.026 & 1.045 & 0.000 \\
IPW & 0.515 & 0.515 & 0.552 & 0.300 & 0.515 & 0.515 & 0.552 & 0.300 & 0.515 & 0.515 & 0.552 & 0.300 & 0.515 & 0.515 & 0.552 & 0.300 \\
OW & 0.184 & 0.184 & 0.208 & 0.567 & 0.184 & 0.184 & 0.208 & 0.567 & 0.184 & 0.184 & 0.208 & 0.567 & 0.184 & 0.184 & 0.208 & 0.567 \\
\addlinespace
\multicolumn{17}{l}{\emph{Parametric proximal bridge \citep{ying2022proximal}}} \\
POR & 0.379 & 0.379 & 0.388 & 0.033 & 0.369 & 0.369 & 0.377 & 0.033 & 0.360 & 0.360 & 0.368 & 0.033 & 0.369 & 0.369 & 0.378 & 0.033 \\
PIPW & 13.468 & 13.468 & 15.173 & 0.652 & 7.625 & 7.625 & 7.766 & 0.000 & 16.214 & 16.214 & 43.099 & 0.926 & 12.435 & 12.435 & 22.012 & 0.526 \\
PDR & 0.498 & 0.498 & 0.511 & 0.033 & 0.463 & 0.463 & 0.470 & 0.000 & 0.368 & 0.368 & 0.373 & 0.000 & 0.443 & 0.443 & 0.451 & 0.011 \\
\addlinespace
\multicolumn{17}{l}{\emph{Flexible proximal bridges}} \\
OB-\modelnameNMMR{} & 0.086 & 0.119 & 0.175 & 0.933 & 0.128 & 0.135 & 0.168 & 0.833 & 0.199 & 0.199 & 0.213 & 0.300 & 0.138 & 0.151 & 0.185 & 0.689 \\
TB-\modelnameNMMR{} & 2.908 & 2.921 & 4.492 & 0.900 & 1.763 & 1.763 & 2.117 & 0.900 & 1.318 & 1.318 & 1.450 & 0.467 & 1.996 & 2.001 & 2.686 & 0.756 \\
DR-\modelnameNMMR{} & -0.543 & 0.548 & 0.588 & 0.333 & -0.032 & 0.113 & 0.159 & 0.933 & 0.149 & 0.151 & 0.209 & 0.833 & -0.142 & 0.271 & 0.319 & 0.700 \\
OB-\modelnamePMMR{} & 1.890 & 1.890 & 1.890 & 0.000 & 0.802 & 0.802 & 0.802 & 0.000 & 0.500 & 0.500 & 0.500 & 0.000 & 1.064 & 1.064 & 1.064 & 0.000 \\
TB-\modelnamePMMR{} & 14.124 & 14.124 & 61.937 & 0.966 & 1.994 & 1.994 & 2.116 & 0.172 & 1.706 & 1.706 & 2.153 & 0.967 & 5.942 & 5.942 & 22.069 & 0.702 \\
DR-\modelnamePMMR{} & 0.810 & 0.810 & 0.858 & 0.133 & 0.817 & 0.817 & 0.820 & 0.000 & 0.485 & 0.485 & 0.487 & 0.000 & 0.704 & 0.704 & 0.722 & 0.044 \\
\bottomrule
\end{tabular}%
}
\end{table}

\begin{table}[t]
\caption{Detailed SEM results for all three sample sizes. Each panel reports Bias, MAE, RMSE, and Coverage at $t\in\{0.25,0.50,0.75\}$ together with their simple averages. Coverage for the proximal bridge estimators is that of pointwise 95\% Wald intervals with influence-function standard errors over 500 replicates, while the SEM Cox baseline coverage values use the replicate-dispersion reconstruction described in Appendix \ref{app:metrics}; the average Bias column reports the average absolute bias.}
\label{tab:sem-detailed}
\centering
% \scriptsize
\resizebox{\textwidth}{!}{%
\begin{tabular}{lcccccccccccccccc}
\toprule
\multicolumn{17}{l}{\textbf{$N=500$}} \\
& \multicolumn{4}{c}{$t=0.25$} & \multicolumn{4}{c}{$t=0.50$} & \multicolumn{4}{c}{$t=0.75$} & \multicolumn{4}{c}{Avg.} \\
Method & Bias & MAE & RMSE & Cov & Bias & MAE & RMSE & Cov & Bias & MAE & RMSE & Cov & Bias & MAE & RMSE & Cov \\
\midrule
Uniform & 0.010 & 0.020 & 0.025 & 1.000 & 0.018 & 0.030 & 0.037 & 1.000 & 0.023 & 0.035 & 0.043 & 1.000 & 0.017 & 0.028 & 0.035 & 1.000 \\
IPW & 0.019 & 0.024 & 0.030 & 1.000 & 0.032 & 0.037 & 0.045 & 1.000 & 0.038 & 0.044 & 0.053 & 1.000 & 0.030 & 0.035 & 0.043 & 1.000 \\
OW & 0.020 & 0.024 & 0.030 & 1.000 & 0.032 & 0.037 & 0.045 & 1.000 & 0.039 & 0.044 & 0.053 & 1.000 & 0.030 & 0.035 & 0.043 & 1.000 \\
POR \citep{ying2022proximal} & -0.000 & 0.021 & 0.026 & 0.034 & -0.001 & 0.031 & 0.039 & 0.068 & -0.001 & 0.035 & 0.045 & 0.128 & 0.001 & 0.029 & 0.037 & 0.077 \\
PIPW \citep{ying2022proximal} & -0.001 & 0.036 & 0.044 & 1.000 & -0.002 & 0.040 & 0.050 & 0.996 & -0.000 & 0.041 & 0.051 & 0.988 & 0.001 & 0.039 & 0.049 & 0.995 \\
PDR \citep{ying2022proximal} & -0.001 & 0.036 & 0.044 & 0.958 & -0.002 & 0.040 & 0.051 & 0.954 & -0.000 & 0.041 & 0.052 & 0.952 & 0.001 & 0.039 & 0.049 & 0.955 \\
OB-\modelnameNMMR{} & 0.022 & 0.037 & 0.046 & 0.076 & 0.015 & 0.041 & 0.053 & 0.072 & 0.043 & 0.058 & 0.070 & 0.040 & 0.027 & 0.046 & 0.056 & 0.063 \\
TB-\modelnameNMMR{} & 0.025 & 0.050 & 0.066 & 0.970 & 0.032 & 0.050 & 0.064 & 0.966 & 0.036 & 0.048 & 0.060 & 0.948 & 0.031 & 0.050 & 0.063 & 0.961 \\
DR-\modelnameNMMR{} & 0.012 & 0.034 & 0.043 & 0.942 & 0.018 & 0.038 & 0.048 & 0.950 & 0.025 & 0.041 & 0.051 & 0.934 & 0.019 & 0.038 & 0.047 & 0.942 \\
OB-\modelnamePMMR{} & 0.001 & 0.032 & 0.040 & 0.450 & -0.002 & 0.039 & 0.048 & 0.438 & 0.001 & 0.040 & 0.049 & 0.418 & 0.001 & 0.037 & 0.046 & 0.435 \\
TB-\modelnamePMMR{} & 0.010 & 0.048 & 0.062 & 0.978 & 0.021 & 0.047 & 0.059 & 0.970 & 0.028 & 0.046 & 0.058 & 0.968 & 0.020 & 0.047 & 0.059 & 0.972 \\
DR-\modelnamePMMR{} & 0.001 & 0.035 & 0.044 & 0.942 & -0.002 & 0.040 & 0.050 & 0.958 & 0.001 & 0.041 & 0.051 & 0.956 & 0.001 & 0.039 & 0.048 & 0.952 \\
\addlinespace
\multicolumn{17}{l}{\textbf{$N=1000$}} \\
& \multicolumn{4}{c}{$t=0.25$} & \multicolumn{4}{c}{$t=0.50$} & \multicolumn{4}{c}{$t=0.75$} & \multicolumn{4}{c}{Avg.} \\
Method & Bias & MAE & RMSE & Cov & Bias & MAE & RMSE & Cov & Bias & MAE & RMSE & Cov & Bias & MAE & RMSE & Cov \\
\midrule
Uniform & 0.010 & 0.015 & 0.018 & 1.000 & 0.017 & 0.023 & 0.028 & 1.000 & 0.022 & 0.027 & 0.033 & 1.000 & 0.016 & 0.022 & 0.027 & 1.000 \\
IPW & 0.019 & 0.021 & 0.024 & 1.000 & 0.031 & 0.033 & 0.038 & 1.000 & 0.037 & 0.039 & 0.045 & 1.000 & 0.029 & 0.031 & 0.036 & 1.000 \\
OW & 0.019 & 0.021 & 0.024 & 1.000 & 0.031 & 0.033 & 0.038 & 1.000 & 0.037 & 0.039 & 0.045 & 1.000 & 0.029 & 0.031 & 0.036 & 1.000 \\
POR \citep{ying2022proximal} & 0.000 & 0.015 & 0.019 & 0.028 & 0.000 & 0.023 & 0.028 & 0.064 & 0.000 & 0.026 & 0.031 & 0.102 & 0.000 & 0.021 & 0.026 & 0.065 \\
PIPW \citep{ying2022proximal} & -0.000 & 0.025 & 0.031 & 1.000 & -0.001 & 0.028 & 0.034 & 0.998 & -0.001 & 0.029 & 0.036 & 0.992 & 0.001 & 0.027 & 0.034 & 0.997 \\
PDR \citep{ying2022proximal} & -0.000 & 0.025 & 0.031 & 0.964 & -0.001 & 0.028 & 0.034 & 0.978 & -0.001 & 0.029 & 0.036 & 0.956 & 0.001 & 0.028 & 0.034 & 0.966 \\
OB-\modelnameNMMR{} & 0.019 & 0.029 & 0.036 & 0.046 & 0.003 & 0.030 & 0.038 & 0.086 & 0.031 & 0.045 & 0.056 & 0.050 & 0.018 & 0.035 & 0.043 & 0.061 \\
TB-\modelnameNMMR{} & 0.016 & 0.044 & 0.055 & 0.954 & 0.026 & 0.041 & 0.051 & 0.946 & 0.030 & 0.041 & 0.051 & 0.916 & 0.024 & 0.042 & 0.052 & 0.939 \\
DR-\modelnameNMMR{} & 0.007 & 0.024 & 0.030 & 0.950 & 0.013 & 0.028 & 0.034 & 0.968 & 0.021 & 0.032 & 0.040 & 0.910 & 0.014 & 0.028 & 0.034 & 0.943 \\
OB-\modelnamePMMR{} & -0.001 & 0.023 & 0.028 & 0.408 & -0.003 & 0.027 & 0.033 & 0.426 & -0.001 & 0.027 & 0.034 & 0.408 & 0.002 & 0.026 & 0.032 & 0.414 \\
TB-\modelnamePMMR{} & 0.010 & 0.046 & 0.057 & 0.942 & 0.021 & 0.041 & 0.052 & 0.936 & 0.027 & 0.039 & 0.049 & 0.918 & 0.019 & 0.042 & 0.053 & 0.932 \\
DR-\modelnamePMMR{} & -0.001 & 0.024 & 0.030 & 0.950 & -0.003 & 0.027 & 0.034 & 0.958 & -0.002 & 0.028 & 0.035 & 0.940 & 0.002 & 0.027 & 0.033 & 0.949 \\
\addlinespace
\multicolumn{17}{l}{\textbf{$N=2000$}} \\
& \multicolumn{4}{c}{$t=0.25$} & \multicolumn{4}{c}{$t=0.50$} & \multicolumn{4}{c}{$t=0.75$} & \multicolumn{4}{c}{Avg.} \\
Method & Bias & MAE & RMSE & Cov & Bias & MAE & RMSE & Cov & Bias & MAE & RMSE & Cov & Bias & MAE & RMSE & Cov \\
\midrule
Uniform & 0.011 & 0.013 & 0.016 & 1.000 & 0.018 & 0.021 & 0.025 & 1.000 & 0.024 & 0.026 & 0.031 & 1.000 & 0.018 & 0.020 & 0.024 & 1.000 \\
IPW & 0.020 & 0.020 & 0.023 & 1.000 & 0.032 & 0.033 & 0.036 & 1.000 & 0.039 & 0.039 & 0.044 & 0.000 & 0.030 & 0.031 & 0.034 & 0.667 \\
OW & 0.020 & 0.021 & 0.023 & 1.000 & 0.032 & 0.033 & 0.037 & 1.000 & 0.039 & 0.039 & 0.044 & 0.000 & 0.030 & 0.031 & 0.035 & 0.667 \\
POR \citep{ying2022proximal} & 0.000 & 0.011 & 0.014 & 0.034 & 0.000 & 0.017 & 0.020 & 0.064 & 0.000 & 0.019 & 0.023 & 0.098 & 0.000 & 0.016 & 0.019 & 0.065 \\
PIPW \citep{ying2022proximal} & 0.001 & 0.018 & 0.023 & 1.000 & -0.001 & 0.022 & 0.027 & 0.994 & -0.001 & 0.021 & 0.026 & 0.984 & 0.001 & 0.020 & 0.025 & 0.993 \\
PDR \citep{ying2022proximal} & 0.001 & 0.018 & 0.023 & 0.940 & -0.001 & 0.022 & 0.027 & 0.954 & -0.001 & 0.022 & 0.026 & 0.956 & 0.001 & 0.020 & 0.025 & 0.950 \\
OB-\modelnameNMMR{} & 0.024 & 0.029 & 0.034 & 0.046 & -0.006 & 0.024 & 0.030 & 0.062 & 0.017 & 0.033 & 0.042 & 0.070 & 0.016 & 0.028 & 0.035 & 0.059 \\
TB-\modelnameNMMR{} & 0.012 & 0.030 & 0.037 & 0.950 & 0.022 & 0.031 & 0.039 & 0.928 & 0.027 & 0.032 & 0.040 & 0.858 & 0.020 & 0.031 & 0.039 & 0.912 \\
DR-\modelnameNMMR{} & 0.008 & 0.018 & 0.023 & 0.922 & 0.012 & 0.022 & 0.027 & 0.920 & 0.018 & 0.025 & 0.031 & 0.878 & 0.012 & 0.022 & 0.027 & 0.907 \\
OB-\modelnamePMMR{} & 0.001 & 0.017 & 0.022 & 0.384 & -0.001 & 0.021 & 0.026 & 0.334 & -0.001 & 0.021 & 0.026 & 0.350 & 0.001 & 0.020 & 0.025 & 0.356 \\
TB-\modelnamePMMR{} & 0.006 & 0.030 & 0.038 & 0.952 & 0.018 & 0.030 & 0.037 & 0.934 & 0.023 & 0.030 & 0.038 & 0.886 & 0.016 & 0.030 & 0.038 & 0.924 \\
DR-\modelnamePMMR{} & 0.001 & 0.019 & 0.023 & 0.932 & -0.002 & 0.022 & 0.027 & 0.938 & -0.002 & 0.021 & 0.027 & 0.946 & 0.002 & 0.021 & 0.026 & 0.939 \\
\bottomrule
\end{tabular}%
}
\end{table}

\begin{table}[t]
\caption{Standard versus H\'ajek-normalized weighted estimators. The rows labeled \citet{ying2022proximal} correspond to the parametric proximal bridge. ACTG averages are over $\{201,401,602,802\}$ days and Framingham averages are over $\{2184,3640,5095\}$ days. Coverage is replicate-dispersion coverage; for Framingham, the target is the external benchmark; see Appendix~\ref{app:metrics}. Framingham PIPW (Standard) excludes the runs whose hazard ratio is not finite (Table~\ref{tab:framingham-detailed}).}
\label{tab:hajek-ablation-full}
\centering
% \scriptsize
\begin{tabular}{llllcccc}
\toprule
Dataset & Bridge & Estimator & Variant & Avg.\ abs.\ bias & MAE & RMSE & Coverage \\
\midrule
ACTG & \citet{ying2022proximal} & PIPW & Standard & 0.025 & 0.033 & 0.038 & 0.98 \\
ACTG & \citet{ying2022proximal} & PIPW & H\'ajek & 0.008 & 0.017 & 0.020 & 1.00 \\
ACTG & \citet{ying2022proximal} & PDR & Standard & 0.015 & 0.022 & 0.025 & 0.97 \\
ACTG & \citet{ying2022proximal} & PDR & H\'ajek & 0.008 & 0.017 & 0.020 & 1.00 \\
ACTG & \modelnameNMMR{} & TB & Standard & 0.013 & 0.027 & 0.036 & 0.98 \\
ACTG & \modelnameNMMR{} & TB & H\'ajek & 0.005 & 0.014 & 0.018 & 1.00 \\
ACTG & \modelnameNMMR{} & DR & Standard & 0.006 & 0.016 & 0.020 & 0.99 \\
ACTG & \modelnameNMMR{} & DR & H\'ajek & 0.004 & 0.015 & 0.019 & 1.00 \\
\addlinespace
Framingham & \citet{ying2022proximal} & PIPW & Standard & 12.435 & 12.435 & 22.012 & 0.53 \\
Framingham & \citet{ying2022proximal} & PIPW & H\'ajek & 0.633 & 0.633 & 0.634 & 0.00 \\
Framingham & \citet{ying2022proximal} & PDR & Standard & 0.443 & 0.443 & 0.451 & 0.01 \\
Framingham & \citet{ying2022proximal} & PDR & H\'ajek & 0.175 & 0.176 & 0.189 & 0.27 \\
Framingham & \modelnameNMMR{} & TB & Standard & 1.996 & 2.001 & 2.686 & 0.76 \\
Framingham & \modelnameNMMR{} & TB & H\'ajek & 0.802 & 0.802 & 0.804 & 0.00 \\
Framingham & \modelnameNMMR{} & DR & Standard & 0.241 & 0.271 & 0.319 & 0.70 \\
Framingham & \modelnameNMMR{} & DR & H\'ajek & 0.269 & 0.290 & 0.329 & 0.66 \\
\bottomrule
\end{tabular}
\end{table}

\section{Additional robustness analyses}
\label{app:robustness}
This appendix details the additional studies summarized in Section~\ref{sec:experiments}. Each uses 30 replicates per setting; average absolute bias is computed over the reported horizons, SEM coverage is Wald coverage, and ACTG coverage is replicate-dispersion coverage (Appendix~\ref{app:metrics}).

\paragraph{Modern causal survival baselines.}
We compare with survival TMLE \citep{cai2020one}, IPCW-AIPW \citep{bai2013doubly}, causal survival forests \citep{cui2023estimating}, and DeepSurv \citep{katzman2018deepsurv} with inverse propensity weighting (Table~\ref{tab:modern-baselines}). These flexible estimators are given the same observed variables as our methods but assume no unmeasured confounding, which fails by construction; they therefore remain biased, and their SEM coverage is poor (0.06--0.66). Restricting them to $X$ alone raises their SEM bias to 0.038--0.056.

\begin{table}[!htbp]
\centering
% \small
\caption{Average absolute bias of modern causal survival baselines and our DR estimators (SEM with $N=2000$; ACTG with $N=2139$).}
\label{tab:modern-baselines}
\begin{tabular}{lcc}
\toprule
Method & SEM & ACTG \\
\midrule
Survival TMLE & 0.0191 & 0.0137 \\
IPCW-AIPW & 0.0256 & 0.0058 \\
Causal survival forest & 0.0256 & 0.0057 \\
DeepSurv + IPW & 0.0312 & 0.0858 \\
DR-\modelnamePMMR{} & 0.0017 & 0.0021 \\
DR-\modelnameNMMR{} & 0.0124 & 0.0058 \\
\bottomrule
\end{tabular}
\end{table}

\paragraph{Weak and invalid proxies (SEM).}
The SEM design (Appendix~\ref{app:simulation-details}) draws $Z\sim\operatorname{Normal}(-0.2-0.3X+0.65U,s_Z^2)$ and $W\sim\operatorname{Normal}(-0.6+0.4X+0.65U,s_W^2)$ with $s_Z=s_W=0.5$. To weaken the proxies, we inflate $s_Z$ and $s_W$ while keeping every structural coefficient fixed, so the confounding to be removed is unchanged and the unadjusted Kaplan--Meier contrast keeps an average absolute bias of 0.0309. Because the proxies have different $X$-loadings (baseline correlations with $U$ of 0.645 and 0.623), each noise standard deviation is calibrated so that its correlation with $U$ is a given fraction of its own baseline. To make $Z$ invalid, we add a direct $Z$-to-outcome effect by replacing the event rate with $0.1+0.6A+0.25X+0.5U+\delta|Z|$, leaving the proxy and treatment equations unchanged, and recompute the truth at each $\delta$ by Monte Carlo with $10^7$ draws. The proxy-relevance statistic is the canonical correlation between $W$ and $Z$ after residualizing both on $X$ (Table~\ref{tab:proxy-stress}).

\begin{table}[!htbp]
\centering
% \small
\caption{DR-\modelnamePMMR{} under weakened (top) and invalid (bottom) proxies on SEM ($N=2000$).}
\label{tab:proxy-stress}
\begin{tabular}{lccccc}
\toprule
$\mathrm{corr}(U,\text{proxy})$ as a fraction of baseline & 1.00 & 0.75 & 0.50 & 0.25 & 0.10 \\
Noise SD $s_Z/s_W$ & 0.50/0.50 & 0.80/0.82 & 1.33/1.37 & 2.79/2.89 & 7.08/7.32 \\
Proxy-relevance statistic & 0.45 & 0.23 & 0.10 & 0.03 & 0.03 \\
Avg.\ abs.\ bias & 0.0017 & 0.0021 & 0.0073 & 0.0329 & 0.0471 \\
Coverage & 0.94 & 0.93 & 0.87 & 0.70 & 0.57 \\
\midrule
Invalidity $\delta$ & 0 & 0.10 & 0.25 & 0.50 & \\
Proxy-relevance statistic & 0.45 & 0.45 & 0.45 & 0.46 & \\
Unadjusted KM contrast, avg.\ abs.\ bias & 0.0309 & 0.0314 & 0.0320 & 0.0322 & \\
Avg.\ abs.\ bias & 0.0017 & 0.0040 & 0.0074 & 0.0128 & \\
Coverage & 0.94 & 0.93 & 0.92 & 0.90 & \\
\bottomrule
\end{tabular}
\end{table}

\paragraph{Censoring intensity and censoring model (ACTG).}
We replace $\log(1000)$ in \eqref{eq:actg-censor} by $\log(c)$ with $c\in\{1000,640,415\}$, which rescales the censoring times by $c/1000$ (apart from the floor at 1) and leaves the rest of the generator unchanged, giving about 25\%, 41\%, and 60\% censoring (Table~\ref{tab:censoring-stress}). The bias of the DR estimators stays in the same range, while the MAE of DR-\modelnamePMMR{} grows from 0.0136 to 0.0177 as fewer events are observed. With 25\% censoring, replacing the Cox censoring model on $(A,W,Z,X)$ used for the IPCW weights by a Cox model on $(A,X)$ or by a covariate-free Kaplan--Meier estimate gives DR-\modelnamePMMR{} average absolute bias 0.0021, 0.0020, and 0.0019 (coverage 1.00 throughout). The main ACTG results already use a Cox censoring model although the true censoring times are log-normal given the covariates. These conclusions are limited to the censoring mechanisms studied here.

\begin{table}[!htbp]
\centering
% \small
\caption{DR estimators on ACTG under increasing censoring (average absolute bias / replicate-dispersion coverage).}
\label{tab:censoring-stress}
\begin{tabular}{lccc}
\toprule
Censoring rate & 25\% & 41\% & 60\% \\
\midrule
DR-\modelnamePMMR{} & 0.0021 / 1.00 & 0.0036 / 1.00 & 0.0026 / 0.99 \\
DR-\modelnameNMMR{} & 0.0058 / 0.99 & 0.0065 / 0.99 & 0.0071 / 1.00 \\
\bottomrule
\end{tabular}
\end{table}

\paragraph{Uncensored ACTG.}
We also remove censoring from the same ACTG generator, so that all events are observed and the IPCW weights equal one (Table~\ref{tab:actg-uncensored}). Without censoring, POR's average absolute bias falls from 0.156 to 0.010, between the two flexible outcome bridges, whereas theirs changes by less than 0.01; the parametric bridge's failure on ACTG therefore arises mainly under censoring. Both DR estimators have replicate-dispersion coverage 1.00 in both.

\begin{table}[!htbp]
\centering
% \small
\caption{Average absolute bias on ACTG with and without censoring.}
\label{tab:actg-uncensored}
\begin{tabular}{lcc}
\toprule
Method & 25\% censored & Uncensored \\
\midrule
OB-\modelnamePMMR{} & 0.0096 & 0.0044 \\
OB-\modelnameNMMR{} & 0.0235 & 0.0137 \\
POR \citep{ying2022proximal} & 0.1562 & 0.0104 \\
DR-\modelnamePMMR{} & 0.0021 & 0.0006 \\
DR-\modelnameNMMR{} & 0.0058 & 0.0003 \\
\bottomrule
\end{tabular}
\end{table}

\paragraph{Weight and proxy diagnostics.}
The effective sample size of the learned treatment-bridge weights, $(\sum_i\hat q_i)^2/\sum_i\hat q_i^2$ \citep{kish1965survey}, reported as a percentage of $n$, is 96.8\% on ACTG and 95.6--99.1\% across SEM conditions but 51.8\% on Framingham, where the coefficient of variation of the weights is 0.97, compared with 0.09--0.21 on SEM and ACTG. The proxy-relevance statistic falls from 0.45 to 0.03 as the SEM proxies are weakened, so weak proxies are visible before any bridge is fit, but it does not change under the invalid-proxy perturbation.

\paragraph{Monotonicity.}
The raw estimators do not constrain the estimated survival curves to be non-increasing, as with most curve-wise causal estimators. Monotone rearrangement \citep{chernozhukov2009improving} changes 0.1--0.4\% of the outcome-bridge estimates, by essentially zero, and 47--83\% of the weighted estimates, by $1.4\times10^{-4}$ on average and at most $1.3\times10^{-3}$. Rearrangement does not increase the estimation error for a monotone target \citep{chernozhukov2009improving}, and it leaves every reported metric unchanged at the printed precision.

\section{Additional implementation details}
\label{app:implementation}
Anonymized code for \modelnamePMMR{} and \modelnameNMMR{}, with the SEM and ACTG data generators and the Framingham pipeline (without the data), is provided in the supplementary material. All experiments were run on an internal computing cluster, with each run of the neural and kernel bridges using a single NVIDIA H200 GPU.

\begin{table}[t]
\centering
\caption{Neural and kernel architectures used by the proposed survival bridge estimators. Here $d_W$, $d_Z$, and $d_X$ denote the dimensions of the outcome proxy, treatment proxy, and observed covariates.}
\label{tab:architectures}
% \small
% \begin{tabular}{p{3.1cm} p{9.3cm}}
\begin{tabular}{p{5cm} p{11cm}}
\toprule
\textbf{Component} & \textbf{Architecture / parameterization} \\
\midrule
\modelnameNMMR{} outcome bridge (ACTG, Framingham) &
Input$(t,a,w,x)\in\mathbb{R}^{2+d_W+d_X}$; FC$(2+d_W+d_X,32)$, ReLU; FC$(32,16)$, ReLU; FC$(16,1)$; Softplus output. \\
\modelnameNMMR{} outcome bridge (SEM) &
Input$(t,a,w,x)\in\mathbb{R}^{2+d_W+d_X}$; FC$(2+d_W+d_X,16)$, ReLU; FC$(16,16)$, ReLU; FC$(16,1)$; Softplus output. \\
Targeted $q_\phi$ for ACTG &
Signed input $\tilde d(a,z,x)=(-1)^{1-a}(1,a,z^\top,x^\top)^\top$; FC$(2+d_Z+d_X,32)$, ReLU; FC$(32,16)$, ReLU; FC$(16,1)$; positive output $1+\exp(\cdot)$. \\
Targeted $q_\phi$ for Framingham &
Signed input $\tilde d(a,z,x)$; FC$(2+d_Z+d_X,32)$, ReLU; FC$(32,32)$, ReLU; FC$(32,1)$; positive output $1+\exp(\cdot)$. \\
Targeted $q_\phi$ for SEM &
Signed input $\tilde d(a,z,x)$; FC$(2+d_Z+d_X,8)$, ReLU; FC$(8,1)$; positive output $1+\exp(\cdot)$. \\
\modelnamePMMR{} outcome bridge &
Treatment-matched kernel on $A$, RBF kernels on standardized $(W,X)$ and $(A,Z,X)$, $128$ sample support points, $48$ time support points, and ridge regularization $\lambda=10^{-2}$. \\
\bottomrule
\end{tabular}
\end{table}

For the Framingham dataset, \modelnameNMMR{} uses a 200-epoch AdamW budget for both the targeted $q$ and neural $h$ models, batch sizes $32$ and $128$, learning rate $10^{-3}$, weight decay $10^{-4}$, validation fraction $0.2$, patience $20$, and 48 time centers; under \modelnamePMMR{}, the outcome bridge $h$ instead uses 128 sample support points and 48 time support points and is fit by a direct regularized kernel solve, while the targeted $q$ bridge is still trained by mini-batch AdamW. For Framingham, Ying and \modelnameNMMR{}/\modelnamePMMR{} report hazard ratios aggregated over 30 runs. We set $\lambda_{\mathrm{norm}}=1.0$ and do not enable adaptive normalization. For the proximal semi-synthetic ACTG dataset, \modelnameNMMR{} uses 1000 AdamW epochs, $h$ batch size $64$, $q$ and $q_{\mathrm{ref}}$ batch size $1000$, learning rate $10^{-2}$, weight decay $5\times 10^{-6}$, 48 time centers, targeted $q$ hidden widths $32,16$, and adaptive normalization with $L=0.25$; under \modelnamePMMR{}, the outcome bridge $h$ again uses 128 sample support points and 48 time support points with a direct regularized kernel solve, while the targeted $q$ bridge remains AdamW-trained. For the ACTG dataset, we report metrics aggregated over 30 semi-synthetic datasets and report the ACTG horizons $\{201,401,602,802\}$ days. For the SEM dataset, \modelnameNMMR{} uses 1000 AdamW epochs, $h$ batch size $200$, $q$ batch size $128$, learning rate $10^{-2}$, weight decay $5\times 10^{-6}$, and 48 time centers, while under \modelnamePMMR{} the outcome bridge $h$ again uses 128 sample support points and 48 time support points with a direct regularized kernel solve; adaptive normalization with $L=0.25$ is used in the targeted $q$ objective, and the targeted $q$ bridge remains mini-batch AdamW optimizer trained.
In the SEM results for each reported sample size $N\in\{500,1000,2000\}$, we run with 500 simulations. All the experiments were run using a single NVIDIA H200 GPU.

\clearpage

\end{document}